\documentclass[10pt,twocolumn,letterpaper]{article}
\usepackage{animate}
\usepackage{graphicx}
\usepackage{subcaption}
\usepackage{adjustbox}
\usepackage{array}
\usepackage{cvpr}
\usepackage{multirow}
\usepackage{minibox}   
\usepackage[most]{tcolorbox}   
\usepackage{calc}      
\usepackage{booktabs}
\usepackage{multirow}
\usepackage{siunitx}
\usepackage[table]{xcolor} 
\usepackage{float}
\definecolor{cvprblue}{rgb}{0.21,0.49,0.74}
\usepackage[pagebackref,breaklinks,colorlinks]{hyperref}
\usepackage{flushend}

\def\confName{CVPR}
\def\confYear{2026}
\definecolor{mygray}{RGB}{240,240,240}

\newcommand{\abbr}{\textbf{Refaçade}} 
\title{Refaçade: Editing Object with Given Reference Texture}

\author{
Youze Huang$^{1,*}$ \quad 
Penghui Ruan$^{2,*}$ \quad 
Bojia Zi$^{3, *}$ \quad 
Xianbiao Qi$^{4, \dagger}$\\
{Jianan Wang$^{5}$} \quad
{Rong Xiao$^{4}$} \\
{\small $^1$University of Electronic Science and Technology of China} \quad 
{\small $^2$The Hong Kong Polytechnic University} \quad\\
{\small $^3$The Chinese University of Hong Kong} \quad 
{\small $^4$IntelliFusion Inc.} \quad
{\small $^5$Astribot Inc.}
}
\usepackage[utf8]{inputenc}
\usepackage{textgreek}
\usepackage{amssymb}

\begin{document}

\twocolumn[{
\renewcommand\twocolumn[1][]{#1}
\maketitle
\begin{center}
  \vspace*{-1em}
  \noindent\makebox[\textwidth][c]{
    \animategraphics[width=1\textwidth]{12}{images/video_teaser/}{1}{24}%
  }
  \vspace*{-0.0em}
  \parbox{0.94\textwidth}{
    \centering
    \captionof{figure}{Visual results of Refaçade on videos. \emph{Best viewed with Adobe Acrobat Reader; click to play.}}
    \label{fig:teaser_video}
  }
  \vspace*{-0.0em}
\end{center}
}]

\begingroup
\renewcommand\thefootnote{}
\footnotetext{$^*$ Equal contribution, $^{\dagger}$ Corresponding author.}%
\addtocounter{footnote}{0}
\endgroup

\begin{figure*}[!htp]
  \centering
  \vspace*{-0.5em}
  \includegraphics[width=1\textwidth]{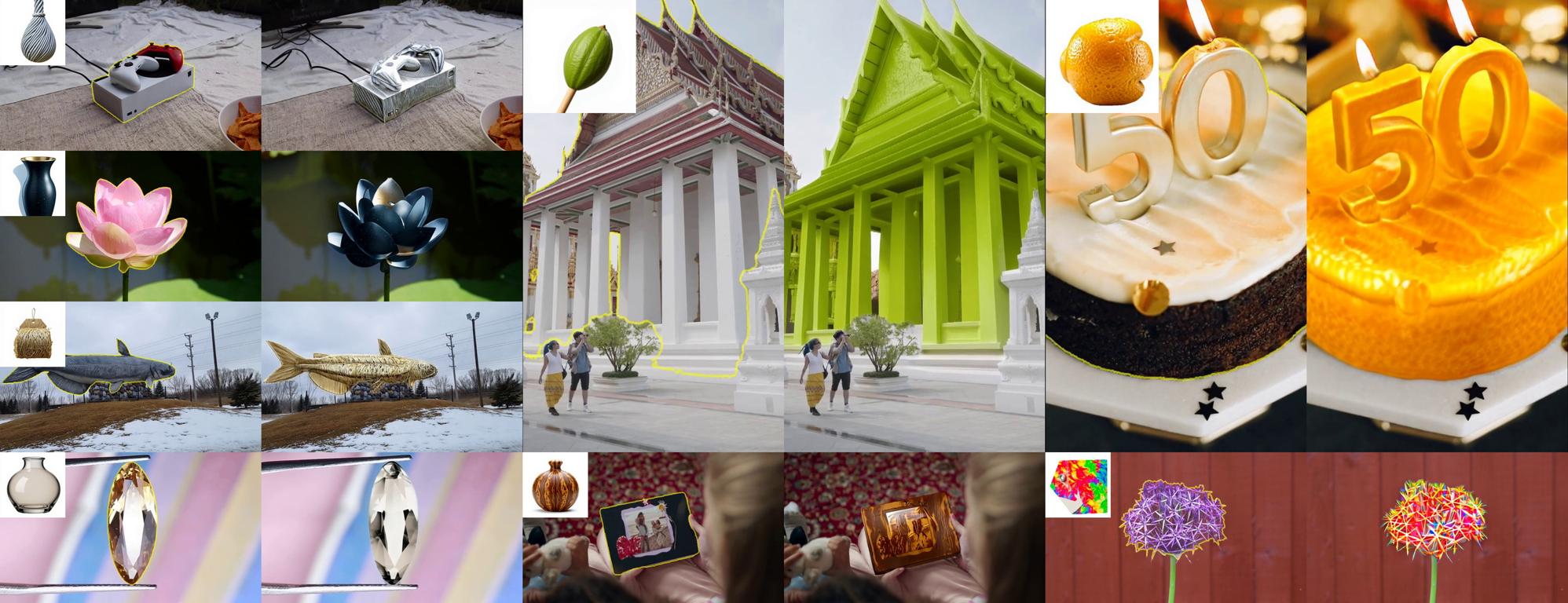}
  \vspace*{-0.5em}
  \caption{
      Visual results of Refaçade on images.}
  \label{fig:image_results}
\end{figure*}

\begin{abstract}
    Recent advances in diffusion models have brought remarkable progress in image and video editing, yet some tasks remain underexplored. In this paper, we introduce a new task, \emph{Object Retexture}, which transfers local textures from a reference object to a target object in images or videos. To perform this task, a straightforward solution is to use ControlNet conditioned on the source structure and the reference texture. However, this approach suffers from limited controllability due to two reasons: 
    conditioning on the raw reference image introduces unwanted structural information, and this method fails to disentangle visual texture and structure information of the source. To address this problem, we proposed a method, namely \textbf{Refaçade}, that consists of two key designs to achieve precise and controllable texture transfer in both images and videos. First, we employ a texture remover trained on paired textured/untextured 3D mesh renderings to remove appearance information while preserving geometry and motion of source videos. Second, we disrupt the reference’s global layout using a jigsaw permutation, encouraging the model to focus on local texture statistics rather than global layout of object. Extensive experiments demonstrate superior visual quality, precise editing, and controllability, outperforming strong baselines in both quantitative and human evaluations. Code is available at \url{https://github.com/fishZe233/Refacade}.

\end{abstract}

\section{Introduction}
In recent years, diffusion models~\citep{stablediffusion_blattmann2023align, flux, animatediff_guo2023animatediff, chen2024videocrafter2, gen3, sora, cogvideox_yang2024cogvideox, hunyuanvideo_kong2024hunyuanvideo, mochi_1, stepvideo_ma2025step, wan2025, powerpaint_zhuang2024task, omniedit_wei2024omniedit, ultraedit_zhao2024ultraedit, hq_edit_hui2024hq, qwenimage_wu2025qwenimagetechnicalreport, cai2025hidream, kolors, pixart_chen2023} have driven remarkable progress in image and video generation. In early stage, UNet-based architectures, exemplified by models such as StableDiffusionV1.5~\citep{stablediffusion_blattmann2023align} and AnimateDiff~\citep{animatediff_guo2023animatediff} have demonstrated impressive capabilities in generating high-quality images and videos. More recently, the field has advanced with the introduction of transformer-based architectures, as seen in groundbreaking works such as flux~\citep{flux}, Qwen-Image~\citep{qwenimage_wu2025qwenimagetechnicalreport}, Sora~\citep{sora}, HunyuanVideo~\citep{hunyuanvideo_kong2024hunyuanvideo} and Wan2.1~\citep{wan2025}, which employ DiT-based structures~\citep{dit_peebles2022scalable} to achieve unprecedented generation quality.

Parallel to these advancements, diffusion-based editing techniques~\citep{instructpix2pix_brooks2023instructpix2pix, magic_brush_zhang2024magicbrush, imageneditor_wang2023imagen, omni_citation_geng2024instructdiffusion, step1x_edit_liu2025step1x_edit, wang2024videocomposer, bian2025videopainter, vivid_10m_hu2024vivid, senorita_zi2025se, video_p2p_liu2024video, tuneavideo_wu2023tune, tokenflow_geyer2023tokenflow, cong2023flatten, ku2024anyv2v, senorita_zi2025se, minimax_remover_zi2025minimax, ju2025editverse, rorem_li2025rorem, objectclear_zhao2025objectclear, avid_zhang2023avid, cococo_zi2024cococo, diffueraser_li2025diffueraser, mtv_inpaint_yang2025mtv,smartbrush_xie2023smartbrush,powerpaint_zhuang2024task,turbofill_xie2025turbofill, attentive_eraser_sun2025attentive, designedit_Jia_Cheng_Yuan_Wang_Li_Jia_Zhang_2025, stablediffusion_blattmann2023align, flux, sdxl_podell2023sdxl, vipaint_agarwal2024vipaintimageinpaintingpretrained, imagen_saharia2022photorealistic, floed_gu2024advanced, ff_vdi_lee2025_ff_vdi_video} have also seen significant progress. However, some editing tasks remain insufficiently explored. In this study, we propose a novel editing task termed \textbf{Object Retexture}, which aims to transfer texture patterns from a reference image onto a target object within a video while preserving the target's geometric structure and leaving surrounding regions unmodified. The fundamental challenge of this task lies in the disentanglement of two key visual components: \textit{texture} (surface patterns, colors, and material properties) and \textit{structure} (shape, geometry, and spatial layout). Specifically, Object Retexture requires: (1) decoupling texture information from the reference image while discarding its structural characteristics; (2) decoupling the target object's structure from the input video while allowing its texture to be modified; and (3) recombining the reference texture with the target structure to generate coherent edited results. This explicit separation ensures that only surface appearance is transferred from the reference, while the target object's geometric details remain intact.

\begin{figure*}[htp]
    \centering
    \includegraphics[width=0.995\linewidth]{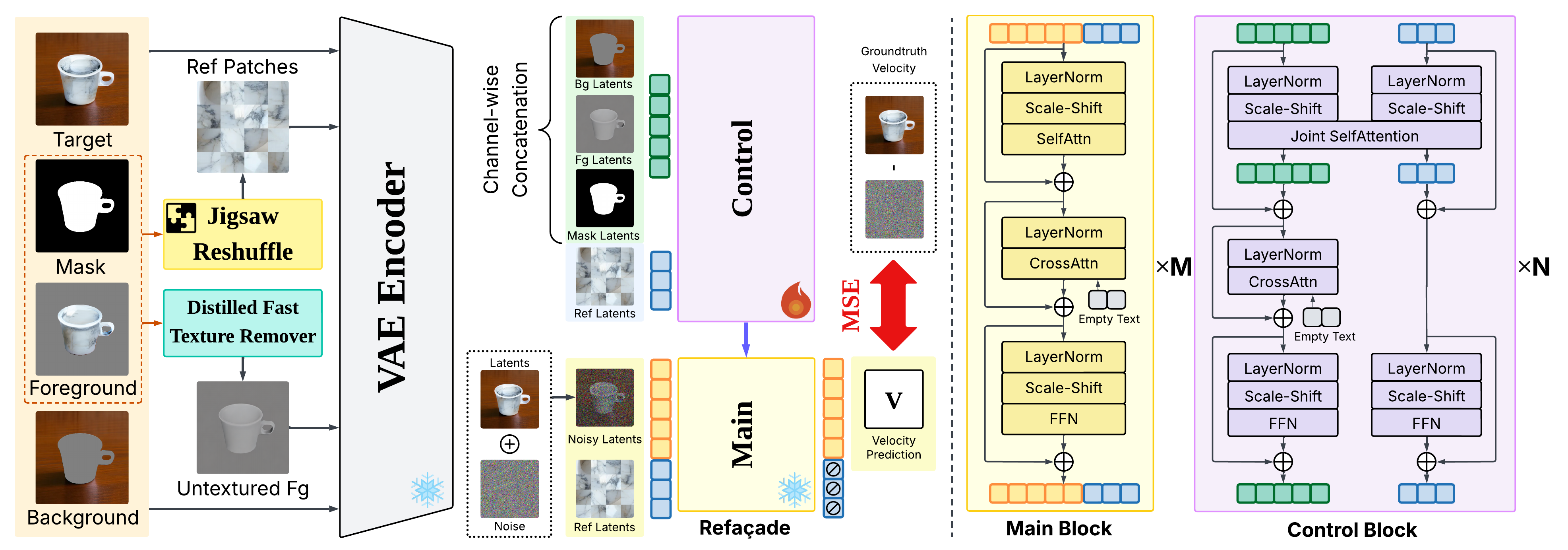}
    \caption{The framework of our Refaçade. The training pipeline of Refaçade is shown on the left, and the model architecture is presented on the right.}
    \label{fig:magic_texture}
\end{figure*}

Nevertheless, Object Retexture can be positioned as a specialized subtask within the broader domain of appearance editing. A natural solution is to condition the control conditions (e.g., HED edges or Canny edges) via ControlNet~\citep{controlnet_zhang2023adding} from source videos to preserve structure, while utilizing the reference image to provide texture information. However, we find this approach fundamentally unsuitable for Object Retexture due to two critical limitations in disentanglement: First, \textit{traditional control signals fail to fully decouple texture from structure}. Conventional control conditions such as depth maps, edge maps, or normal maps are designed to capture geometric information, yet they inevitably retain residual texture cues—such as surface patterns, material boundaries, or color gradients—that should be modifiable rather than preserved. This incomplete disentanglement prevents clean separation between what should be retained (target structure) and what should be modified (target texture). Second, \textit{directly conditioning on the raw reference image introduces unwanted structural information}. When the entire reference image is used as a conditioning signal without proper decoupling, the model inadvertently transfers not only the desired texture patterns but also the reference's geometric characteristics, such as object shape, pose, and spatial layout. This structural leakage from the reference contaminates the target object, resulting in unintended deformations that violate the core requirement of preserving the target's original geometry.

To address these limitations, we propose \abbr{}, a novel framework designed to enhance controllability and suppress unwanted information during texture transfer. Our method comprises two key components. \textit{First, we replace traditional control conditions with texture-free representations rendered from 3D object meshes, which preserve the structural information of the original object while excluding color and texture cues.} To avoid the computational overhead of 3D construction and rendering, we train a texture-remover that directly eliminates texture in the image/video space, eliminating the computational burden associated with 3D lifting and 2D reprojection operations. To achieve fast and accurate texture removal for both images and videos, we train a generator based on Wan2.1~\citep{wan2025} and further distill it using DMD2~\citep{dmd2_yin2024dmd2}, reducing the sampling steps from 50 to just 3. \textit{Second, we introduce a jigsaw permutation strategy that shuffles the reference image to disrupt its spatial structure.} This forces the model to concentrate on the texture itself rather than the object’s shape, effectively preventing the transfer of undesired structural information to the edited object. By combining these two strategies, our approach completely removes the original texture of the source object and ensures that the retextured results are guided solely by the reference texture. Consequently, \abbr{} can accurately edit the appearance of the target object according to the reference texture while preserving the surrounding regions unchanged.

Our main contributions are summarized as follows:
\begin{itemize}[leftmargin=*,itemsep=0.1em]
    \item We introduce a new task, Object Retexture, which enables users to edit an object by refering the texture from a reference image. This task eliminates the need for ambiguous texture prompt when editing an object, allowing users to directly transfer the reference texture onto the source object while preserving the object’s original structure.
    
    \item We propose \abbr{}, an unified model for object retexture in image and video. It consists of two strategies to enhance the controllability of texture transfer and reduce the interference of unwanted information. First, we train a generator to convert objects into texture-free representations, replacing the traditional condition extractor. Second, we apply a jigsaw permutation to disrupt the spatial shape of the object in reference image, encouraging the model to focus more on the texture itself.
    
    \item We conduct extensive experiments across multiple benchmarks, demonstrating that our method achieves superior performance in object retexturing, producing more precise editing results, higher similarity between the reference and edited textures, and better preservation of the surrounding regions.
\end{itemize}

\section{Methodology}

\abbr{} employs two key decoupling strategies, as illustrated in Figure~\ref{fig:magic_texture}: \textbf{Texture Remover} (Sec.\ref{sec:texture_remover}) uses a dedicated diffusion model to remove all texture information from source videos, producing geometry-only representations; and \textbf{Jigsaw Permutation} (Sec.\ref{sec:jigsaw}) applies an effective permutation strategy to remove structural information from the reference image while preserving its texture. 

Given a source video \(\mathbf{X}\), its corresponding object mask \(\textbf{M}\), background video \(\mathbf{X}^{\text{bg}}\), and reference image \(\mathbf{I}^{\text{ref}}\), we first apply the texture remover to obtain an untextured video \(\mathbf{X}^{\text{unt}}\), then apply jigsaw permutation to create a structure-agnostic texture guide. Finally, our texture transfer model synthesizes the output by combining geometric structure from the texture-free source with texture patterns from the permuted reference. \abbr{} is trained with flow matching~\citep{flow_match_lipman2023flow}. Let \(\mathbf{z}_0 = \mathcal{E}_{\mathrm{VAE}}(\mathbf{X})\) denote the target latent. The conditioning signal \(\mathbf{c}\) comprises multiple components:
\[
\mathbf{c} = \Big\{\,
\mathcal{E}_{\mathrm{VAE}}\big(\text{Jigsaw}(\mathbf{I}^{\text{ref}})\big),\;
\mathcal{E}_{\mathrm{VAE}}(\mathbf{X}^{\text{unt}}),\;
\mathbf{M},\;
\mathcal{E}_{\mathrm{VAE}}(\mathbf{X}^{\text{bg}})
\,\Big\},
\]
We sample \(t\sim\mathcal{U}(0,1)\) and \(\boldsymbol{\varepsilon}\sim\mathcal{N}(\mathbf{0},\mathbf{I})\) with the same shape as \(\mathbf{z}_0\), and define the linear interpolation path and target velocity:
\[
\mathbf{z}_t = (1-t)\mathbf{z}_0 + t\boldsymbol{\varepsilon},
\qquad
\mathbf{v}^\star(\mathbf{z}_t,t) = \boldsymbol{\varepsilon}-\mathbf{z}_0.
\]
A velocity network $\mathbf{v}_\theta(\mathbf{z}_t,\mathbf{c},t)$ is trained with the flow-matching loss~\citep{flow_match_lipman2023flow}:
\[
\mathbb{E}_{(\mathbf{z}_0,\,\mathbf{c})}
\;\mathbb{E}_{t\sim\mathcal{U}(0,1),\,\boldsymbol{\varepsilon}\sim\mathcal{N}(\mathbf{0},\mathbf{I})}
\!\left[
\left\|\,
\mathbf{v}_{\boldsymbol{\theta}}(\mathbf{z}_t,\mathbf{c},t)
\;-\;
\mathbf{v}^\star(\mathbf{z}_t,t)
\,\right\|_2^2
\right].
\]

\begin{figure*}[htp]
    \centering
    \includegraphics[width=0.985\linewidth]{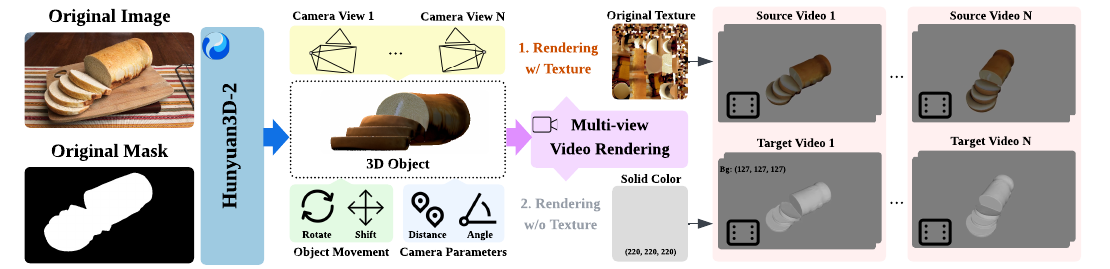}
    \caption{Our data construction pipeline for the texture remover operates as follows: we collect object images, reconstruct 3D meshes, and render paired videos with and without textures under diverse camera trajectories and object motions.}
    \label{fig:texture_remover}
\end{figure*}

Our framework builds upon VACE~\citep{vace_jiang2025vace}, with modifications inspired by MM-DiT~\citep{mmdit_esser2024scaling} to better handle distinct conditioning signals. In the control branch, we concatenate background, texture-free video, and mask latents channel-wise and process them through dedicated condition layers, while reference image latents are processed through separate reference layers. This design allows tokens serving different functions (reference vs. source) to use distinct parameters while sharing the same attention mechanism. In the main branch, the reference image is prepended to the first frame of the noisy latent, and hidden states from the control block are added to corresponding layers.

\subsection{Texture Remover for Source Image and Video}
\label{sec:texture_remover}
In 3D mesh representations, object geometry and texture are inherently decoupled: the mesh defines shape through vertices and faces, while appearance is specified separately via texture coordinates and material properties. A naïve solution would be to reconstruct a 3D object mesh from the video and render it in a texture-free manner to obtain geometry-only conditioning signals. However, classical 3D reconstruction from video is computationally expensive, typically requiring several minutes to recover a textured mesh from a single video clip, making it impractical for large-scale training and inference.
To obtain geometry supervision efficiently at scale, we train a dedicated diffusion model—the \textit{texture remover}—that learns to map textured video frames directly to texture-free frames of the same object. Specifically, we construct a paired training dataset by rendering 3D objects twice: once with full texture maps applied and once with textures removed using uniform gray materials. We then train a video diffusion model to learn this texture removal mapping directly in 2D space, eliminating the need for explicit 3D reconstruction at inference time. Once trained, this model provides efficient geometry-only control signals for arbitrary video clips while preserving object motion, pose, and shape, ensuring precise temporal and spatial alignment with the source video.

\noindent\textbf{Dataset Construction.} The full pipeline is illustrated in Figure~\ref{fig:texture_remover}. We begin by collecting a large-scale image dataset containing commonly observed objects from two sources: (1) first frames extracted from real-world videos, and (2) synthetic images generated via text-to-image models using object-centric prompts (e.g., ``a chair,'' ``a car''). For each image, we segment the main object using an off-the-shelf segmentation model~\citep{sam2_ravi2025sam2} and reconstruct a textured 3D mesh using Hunyuan3D~\cite{zhao2025hunyuan3d}. 

For each reconstructed mesh, we generate paired video sequences as follows. First, we render the mesh with full texture maps applied under fixed camera intrinsics and headlight-style point light while varying camera distance and viewing angle over time. This produces a short video clip capturing the object's original textured appearance. Second, we render the same mesh under identical camera and lighting conditions, but with all texture maps and albedo information removed, using a uniform gray Lambertian material. This geometry-only rendering serves as the texture-free target. To increase dataset diversity and improve model robustness, we apply controlled augmentation by varying: (1) camera trajectories (e.g., orbital, arc, zoom-in/out), (2) light intensity, and (3) object poses (random rotations and translations within reasonable bounds). 

\noindent\textbf{Training and Distillation.} Our texture remover builds on the VACE framework. The input is the source video after background removal, which serves as the control signal. The training objective is to generate the aligned textureless video. We update only the control blocks in VACE while keeping the main branch frozen, thereby restricting learning to the part of the network that translates appearance into geometry. A direct model of this form still requires a large number of denoising steps during sampling, which would significantly increase the total training cost of our full \abbr{} system. To address this issue, we apply DMD~\cite{dmd2_yin2024dmd2} distillation on the trained remover. After distillation, the sampling schedule is reduced from fifty steps to three steps while maintaining the ability to output high-quality texture-free videos.

\subsection{Refaçade: Jigsaw Permutation for Structure-Agnostic Texture Transfer}
\label{sec:jigsaw}

\begin{figure}[htp]
    \centering
    \includegraphics[width=0.985\linewidth]{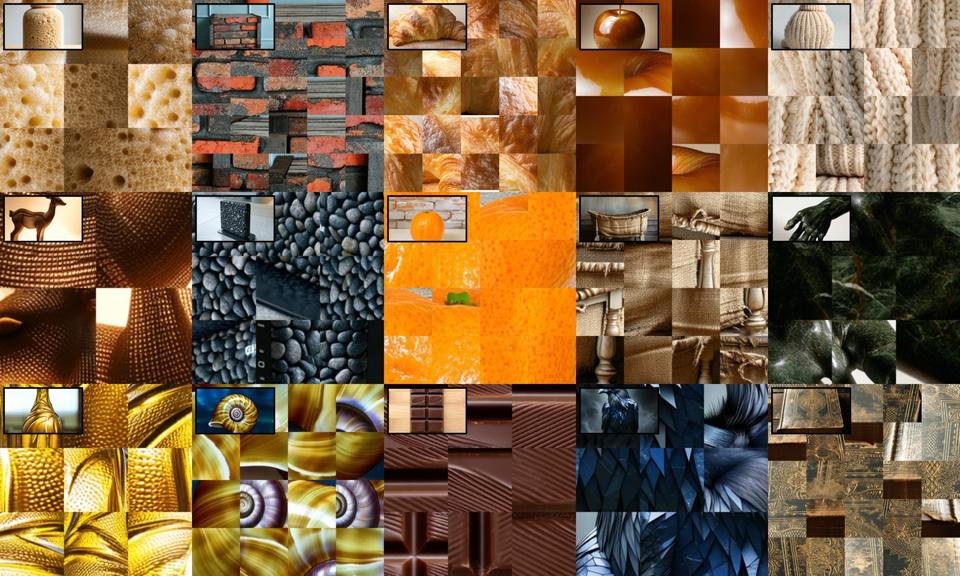}
    \caption{Visualization of Jigsaw Permutation. We extract foreground patches from the reference image on the top-left corner, shuffle and flip them randomly, then rearrange them into a new layout. This destroys global spatial structure while preserving local texture patterns.}
    \label{fig:Jigsaw Patches}
    \vspace{-1em}
\end{figure}

\begin{figure*}[htbp]
  \centering
  \renewcommand{\arraystretch}{1.0}
  \setlength{\tabcolsep}{0pt}

  {
    \fontsize{8pt}{9.6pt}\selectfont
    \begin{tabular}{@{}*{8}{>{\centering\arraybackslash}m{0.125\textwidth}}@{}}
      \minibox{Source} & \minibox{Reference}& \minibox{Mask} & \minibox{BrushNet} & \minibox{ControlNet} &
      \minibox{Flux-Fill}& \minibox{Flux-Kontext-I}  & \minibox{Flux-Kontext-T}  \\
    \end{tabular}
  }

  \vspace{-10pt}

  \begin{subfigure}[b]{\textwidth}
    \centering
    \includegraphics[width=\linewidth]{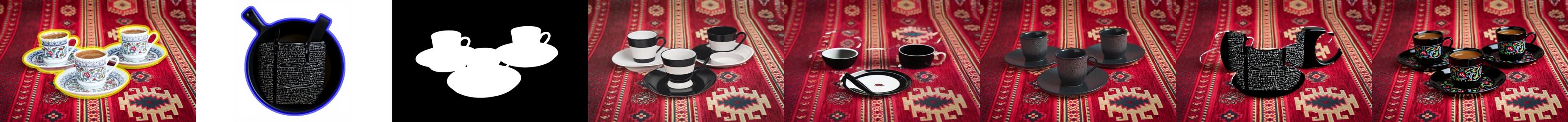}
  \end{subfigure}

  {
    \fontsize{8pt}{9.6pt}\selectfont
    \vspace{-4pt}
    \begin{tabular}{@{}*{8}{>{\centering\arraybackslash}m{0.125\textwidth}}@{}}
      \minibox{HiDream-E1} &\minibox{HQ-Edit} & \minibox{InsP2P} & \minibox{Qwen-Image-Edit} & \minibox{SD3-Inpaint} &
      \minibox{UltraEdit} & \minibox{Nano Banana} & \minibox{Ours} \\
    \end{tabular}
  }

  \vspace{-10pt}

  \begin{subfigure}[b]{\textwidth}
    \centering
    \includegraphics[width=\linewidth]{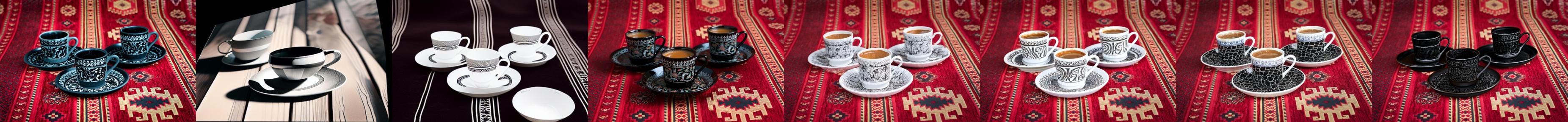}
  \end{subfigure}

  {%
    \fontsize{8pt}{9.6pt}\selectfont
    \begin{tabular}{@{}*{8}{>{\centering\arraybackslash}m{0.125\textwidth}}@{}}
      \minibox{Source} & \minibox{Reference}& \minibox{Mask} & \minibox{BrushNet} & \minibox{ControlNet} &
      \minibox{Flux-Fill}& \minibox{Flux-Kontext-I}  & \minibox{Flux-Kontext-T}  \\
    \end{tabular}
  }

  \vspace{-10pt}

  \begin{subfigure}[b]{\textwidth}
    \centering
    \includegraphics[width=\linewidth]{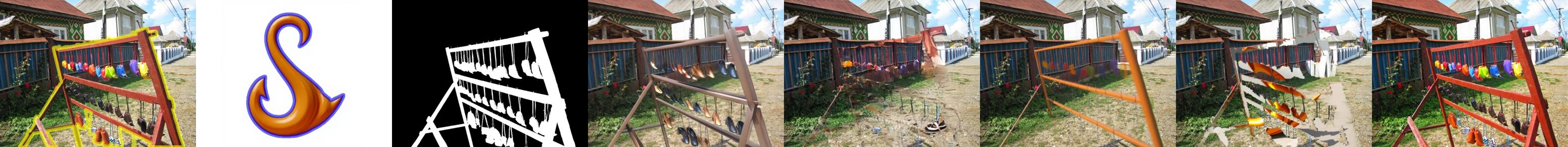}
  \end{subfigure}

  {%
    \fontsize{8pt}{9.6pt}\selectfont
    \vspace{-4pt}
    \begin{tabular}{@{}*{8}{>{\centering\arraybackslash}m{0.125\textwidth}}@{}}
      \minibox{HiDream-E1} &\minibox{HQ-Edit} & \minibox{InsP2P} & \minibox{Qwen-Image-Edit} & \minibox{SD3-Inpaint} &
      \minibox{UltraEdit} & \minibox{Nano Banana} & \minibox{Ours} \\
    \end{tabular}
  }

  \vspace{-10pt}

  \begin{subfigure}[b]{\textwidth}
    \centering
    \includegraphics[width=\linewidth]{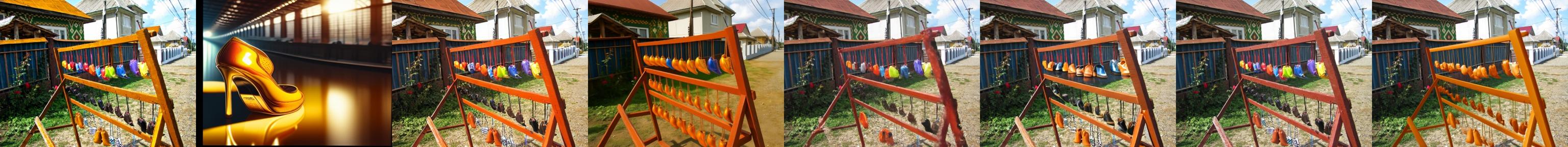}
  \end{subfigure}
  \vspace{-4pt}

  {%
    \fontsize{8pt}{9.6pt}\selectfont 
    \begin{tabular}{@{}*{8}{>{\centering\arraybackslash}m{0.125\linewidth}}@{}}
      \minibox{Source} & \minibox{Reference} & \minibox{Mask} & \minibox{AnyV2V} &
      \minibox{COCOCO} & \minibox{Ditto} & \minibox{Flatten} & \minibox{ICVE} \\
    \end{tabular}%
  }

  \vspace{-1pt}
  \begin{subfigure}{\linewidth}
    \centering
    \animategraphics[width=\linewidth]{12}{images/row1/}{1}{24}
  \end{subfigure}

  {%
    \fontsize{8pt}{9.6pt}\selectfont
    \vspace{-4pt}
    \begin{tabular}{@{}*{8}{>{\centering\arraybackslash}m{0.125\linewidth}}@{}}
      \minibox{InsV2V} & \minibox{InsVIE} & \minibox{Lucy-Edit} & \minibox{Señorita} &
      \minibox{Tokenflow} & \minibox{VACE} & \minibox{VideoPainter} & \minibox{Ours} \\
    \end{tabular}%
  }

  \vspace{-1pt}
  \begin{subfigure}{\linewidth}
    \centering
    \animategraphics[width=\linewidth]{12}{images/row2/}{1}{24}
  \end{subfigure}

  {%
    \fontsize{8pt}{9.6pt}\selectfont 
    \begin{tabular}{@{}*{8}{>{\centering\arraybackslash}m{0.125\linewidth}}@{}}
      \minibox{Source} & \minibox{Reference} & \minibox{Mask} & \minibox{AnyV2V} &
      \minibox{COCOCO} & \minibox{Ditto} & \minibox{Flatten} & \minibox{ICVE} \\
    \end{tabular}%
  }

  \vspace{-1pt}
  \begin{subfigure}{\linewidth}
    \centering
    \animategraphics[width=\linewidth]{12}{images/row3/}{1}{24}
  \end{subfigure}

  {%
    \fontsize{8pt}{9.6pt}\selectfont
    \vspace{-4pt}
    \begin{tabular}{@{}*{8}{>{\centering\arraybackslash}m{0.125\linewidth}}@{}}
      \minibox{InsV2V} & \minibox{InsVIE} & \minibox{Lucy-Edit} & \minibox{Señorita} &
      \minibox{Tokenflow} & \minibox{VACE} & \minibox{VideoPainter} & \minibox{Ours} \\
    \end{tabular}%
  }

  \vspace{-1pt}
  \begin{subfigure}{\linewidth}
    \centering
    \animategraphics[width=\linewidth]{12}{images/row4/}{1}{24}
  \end{subfigure}

  \caption{Comparison results of Refaçade and baselines on both images and videos. First 4 rows: images. Bottom 4 rows: videos. \emph{Best viewed with Adobe Acrobat Reader; click to play.}}
  \label{fig:image_video_compare}
\end{figure*}

While the texture remover ensures that no source appearance leaks through geometric conditioning, we must also prevent the model from copying the reference image's global layout. A straightforward approach would be to use the first frame of the target video (with background removed) as the reference image during training. However, this strategy introduces a critical problem: the reference image and target video would share identical spatial structure, causing the model to learn spatial alignment rather than texture transfer. During inference, when the reference and source objects have different shapes or poses, this approach fails catastrophically---the model attempts to transfer structural characteristics rather than appearance patterns.

To bridge this gap between training and inference, we employ a \textit{Jigsaw Permutation} strategy that forces the model to focus on texture rather than object structure. As illustrated in Figure~\ref{fig:Jigsaw Patches}, we cut square patches from the foreground area of the reference image. To ensure sufficient reference texture within each patch, we discard any patch containing more than 10\% background pixels. We then randomly shuffle and flip these patches, rearranging them into a rectangular area.

\begin{table*}[htbp]
\centering
\scriptsize
\setlength{\tabcolsep}{5pt}

\renewcommand{\arraystretch}{1.25}
\caption{Evaluation on image dataset. The LPIPS for background evaluates background perseveration, while LPIPS for foreground evaluates the similarity between reference texture and generated content. CLIP, DINO and Dream are the abbreviation of CLIPScore, DINOScore and DreamSim, respectively. The best results are \textbf{boldfaced}, and the second-best results are \underline{underlined}.}

\begin{tabular}{
    l|c|cccc|cccc|cc|c
}
\toprule
\multirow{2}{*}{Method}& \multirow{2}{*}{Type}&\multicolumn{4}{c|}{Background} & \multicolumn{4}{c|}{Foreground} & \multicolumn{2}{c|}{LLM Evaluation} & \multirow{2}{*}{User Preference}  \\ \cline{3-12}
& &{MSE$\downarrow$} & {PSNR$\uparrow$} & {SSIM$\uparrow$} & {LPIPS$\downarrow$}
& {CLIP$\uparrow$} & {DINO$\uparrow$} & {LPIPS$\downarrow$} & {Dream$\uparrow$}& {GPT-5$\uparrow$} & {Gemini$\uparrow$}  \\
\toprule
BrushNet ~\citep{brushnet_ju2024brushnetplugandplayimageinpainting}
& \multirow{4}{*}{Inpainting} &438.49 & 23.82 & 0.7555 & 0.1758
& 0.7026 & 0.2235 & 0.7162 & 0.7341
& 2.12& 2.12 & 0.1366\\
ControlNet-Inp ~\cite{ControlNetInpaint}
& &{429.90} & {27.38} & {0.7341} & {0.2386}
& {0.7025} & {0.1840} & {0.7281} & {0.7210}
& {1.41} & {2.12} & 0.1304\\
Flux-Fill ~\cite{flux}
& &{67.05} & {31.92} & {0.8948} & {0.0730}
& {0.6900} & {0.2091} & {0.7431} & {0.7134}
& {2.71} & {1.98} & 0.1615\\
SD3-Inpaint ~\cite{sd3_esser2024scaling}& &{65.66} & {32.35} & {0.8882} & {0.0914} & {0.6617} & {0.1534} & {0.7537} & {0.6821} & 1.29& 1.40 & 0.0311\\ \hline
UltraEdit ~\citep{ultraedit_zhao2024ultraedit} & \multirow{8}{*}{General}&{168.56} & {30.04} & {0.7859} & {0.2216} & {0.6910} & {0.1965} & {0.7373} & {0.7122} &2.16 & 2.16&0.0621\\ 
Flux-Kont-I~\citep{blackforestlabs2025fluxkontext} & & {91.76} & {31.63} & {0.8593} & {0.1321} & \underline{0.7768} & \underline{0.4216} & \underline{0.6607} & \underline{0.8015} & 1.71& 1.65&0.2236\\ 
Flux-Kont-T~\citep{blackforestlabs2025fluxkontext} && {1038.27} & {24.16} & {0.7337} & {0.2126} & {0.6770} & {0.1956} & {0.7131} & {0.7025} & 2.37& 2.52& 0.1553\\ 
HiDream-E1~\citep{hidreami1technicalreport} && {1187.49} & {25.55} & {0.7862} & {0.2403} & {0.6866} & {0.1981} & {0.7140} & {0.7170} & 2.41& 2.38&0.1491\\ 
HQ-Edit ~\citep{hq_edit_hui2024hq} && {8026.55} & {9.74} & {0.4355} & {0.5654} & {0.7046} & {0.2223} & {0.7267} & {0.7305} & 1.56& 0.90& 0.0621\\ 
InsP2P ~\citep{instructpix2pix_brooks2023instructpix2pix} && {2712.53} & {16.58} & {0.6156} & {0.4087} & {0.7035} & {0.2003} & {0.7166} & {0.7292} & 1.92& 1.73&0.1180\\
Qwen-I-Edit~\citep{qwenimage_wu2025qwenimagetechnicalreport} && {1183.89} & {21.84} & {0.6868} & {0.2592} & {0.6868} & {0.2196} & {0.7034} & {0.7161} & \underline{2.78}& 2.76&0.1366\\ 
NanoBanana~\citep{google_nanobanana} && {481.66} & {27.47} & {0.7547} & {0.1446} & {0.6981} & {0.2582} & {0.7247} & {0.7316} & 2.65& 2.41& 0.1553\\ \hline
\textbf{Ours(stage1)} & \multirow{2}{*}{Inpainting} & \underline{49.66} & \underline{36.19} & \textbf{0.8994} & \textbf{0.0472} & 0.7125 & {0.2665} & 0.6915 & 0.7497 & 2.77&\textbf{2.81} & \underline{0.5714}\\
\textbf{Ours(stage2)} & &\textbf{49.36} & \textbf{36.20} & 0.8987 & \underline{0.0487} & \textbf{0.7774} & \textbf{0.4516} & \textbf{0.6181} & \textbf{0.8184} & \textbf{2.89}& \underline{2.77} &\textbf{0.8944}\\\hline
\bottomrule
\end{tabular}

\label{tab:image-table}
\end{table*}

\begin{table*}[htbp]
\centering
\scriptsize
\setlength{\tabcolsep}{4pt}
\renewcommand{\arraystretch}{1.18}
\caption{Evaluation results on video dataset. The LPIPS for background evaluates background perseveration, while LPIPS for foreground evaluates the similarity between reference texture and generated content. CLIP, DINO and Dream are the abbreviation of CLIPScore, DINOScore and DreamSim, respectively. Ewarp is at the range of $1\times 10^{-3}$. The best results are \textbf{boldfaced}, the second-best are \underline{underlined}.}

\begin{tabular}{
    l|c|cccc|cccc|c|cc|c
}
\toprule
\multirow{2}{*}{Method}& \multirow{2}{*}{Type}&\multicolumn{4}{c|}{Background} & \multicolumn{4}{c|}{Foreground} & \multicolumn{1}{c|}{Motion} & \multicolumn{2}{c|}{LLM Evaluation} & \multirow{2}{*}{User Preference}\\ \cline{3-13}
& &{MSE$\downarrow$} & {PSNR$\uparrow$} & {SSIM$\uparrow$} & {LPIPS$\downarrow$}
& {CLIP$\uparrow$} & {DINO$\uparrow$} & {LPIPS$\downarrow$} & {Dream$\uparrow$}
& {EWarp $\downarrow$} & {GPT-5$\uparrow$} & {Gemini$\uparrow$} \\
\midrule
COCOCO ~\citep{cococo_zi2024cococo}
& \multirow{3}{*}{Inpainting}&{164.73} & {29.09} & {0.8259} & {0.1226}
& {0.7125} & {0.1381} & {0.7372} & {0.7286}
& {2.1697} &1.88 & 2.18& 0.1180 \\
VACE ~\citep{vace_jiang2025vace}
& &{1596.84} & {19.73} & {0.7107} & {0.2941}
& {0.7159} & {0.1348} & {0.8009} & {0.7240}
& {1.7763} & 1.90& 2.40 & 0.0497 \\
VideoPainter ~\citep{bian2025videopainter}
& &{64.69} & {32.89} & {0.9072} & {0.1052}
& {0.7130} & {0.1554} & {0.7377} & {0.7173}
& {1.9965} &1.92 & 2.12& 0.0559\\ \hline
AnyV2V ~\citep{ku2024anyv2v}
& \multirow{9}{*}{General}&498.49 & 22.77 & 0.7420 & 0.1983
& 0.7178 & 0.1603 & 0.7382 & 0.7253
& {3.5600} &2.21 & 2.18& 0.0932\\
Ditto ~\cite{ditto_bai2025scaling}
& &{2097.47} & {19.28} & {0.7144} & {0.3084}
& {0.6907} & {0.1229} & {0.8264} & {0.6976}
& {1.3656} &1.20 & 1.20& 0.1366\\
Flatten~\citep{cong2023flatten}
& &{2187.84} & {15.66} & {0.6325} & {0.4308}
& {0.7303} & {0.1708} & {0.7731} & {0.7374}
& {1.7492} & 1.62& 1.38& 0.0745\\
TokenFlow ~\citep{tokenflow_geyer2023tokenflow}
& &{889.93} & {19.73} & {0.7107} & {0.2941}
& {0.7162} & {0.1502} & {0.7884} & {0.7257}
& {1.7625} & 1.69& 1.16& 0.0683\\
ICVE ~\citep{icve_liao2025context}
& &{1703.99} & {19.02} & {0.7095} & {0.3098}
& {0.7198} & {0.1705} & {0.7766} & {0.7359}
& {1.7486} & 2.04& 1.28& 0.1615\\
InsV2V ~\cite{insv2v_cheng2024consistent}
& &{3685.70} & {13.88} & {0.5556} & {0.4733}
& {0.7163} & {0.1389} & {0.7802} & {0.7183}
& {2.7225} & 2.00& 1.83& 0.1429\\
InsVIE ~\citep{wu2025insvie}
& &{5450.47} & {11.94} & {0.4435} & {0.5428}
& {0.7172} & {0.1846} & {0.8145} & {0.7448}
& {3.3529} & 2.12& 1.70& 0.1242\\
Lucy-Edit ~\citep{decartai2025lucyedit}
& &{855.43} & {24.57} & {0.8204} & {0.1653}
& {0.6992} & {0.1463} & {0.7969} & {0.7063}
& {1.5283} & 1.84& 2.23& 0.0683\\
Se\~norita~\citep{senorita_zi2025se}
& &{130.53} & {28.90} & {0.8634} & {0.1754}
& {0.6976} & {0.1503} & {0.7497} & {0.7036}
& {1.3519} & 2.10& 2.34& 0.0621\\ \hline
\textbf{Ours(stage1)}
& \multirow{2}{*}{Inpainting} & \underline{30.66} & \underline{36.44} & \underline{0.9460} & \underline{0.0379}
& \underline{0.7331} & \underline{0.2622} & \underline{0.6540} & \underline{0.7473}
& \textbf{1.3510} & \underline{2.72}& \textbf{3.27}& \underline{0.5155}\\ 
\textbf{Ours(stage2)} &&\textbf{30.35} & \textbf{36.48} & \textbf{0.9485} & \textbf{0.0344} & \textbf{0.7524} & \textbf{0.3241} & \textbf{0.6080} & \textbf{0.7742} & \underline{1.4248} & \textbf{2.82}& \underline{3.25} &\textbf{0.7391}\\ \hline
\bottomrule
\end{tabular}

\label{tab:video-table}
\end{table*}

Crucially, we resize the crafted reference patches to match the canvas width used during training, but allow the height to vary based on the number of patches. This ensures that the reference patches have a different aspect ratio and spatial layout compared to the source object. By training on such spatially-permuted references, the model learns to extract and transfer local texture patterns rather than memorizing global spatial configurations. This facilitates strong generalization: at inference time, the model can successfully transfer textures even when the reference and source objects have vastly different shapes, sizes, or poses.

In the training stage, given a source video or image $\mathbf{X}$, the texture remover module will generate a untextured video or image $\mathbf{X}^{\text{unt}}$. We use jagsaw to permutate the first frame of the source to obtain $\mathbf{I}^{\text{ref}}$. Our final training target is to reconstruct the original video $\mathbf{X}$.

\section{Experiments}

\noindent\textbf{Training Dataset.} We use watermark-free WebVid-10M dataset~\citep{webvid_bain2021frozen} and the Pexels dataset~\citep{pexels}. Object category names are first extracted using CogVLM2~\citep{Cogvlm2_hong2024cogvlm2}, and the corresponding segmentation masks are generated with Grounded-SAM2~\citep{groundingdino_liu2023grounding, sam2_ravi2025sam2}. Only masks with good quality are retained. After filtering, we have approximately 1.8 million videos for WebVid-10M and around 180K for Pexels.

\begin{table*}[htbp]
\centering
\scriptsize
\setlength{\tabcolsep}{4pt}
\renewcommand{\arraystretch}{1.18}
\caption{Ablation study for our training pipeline. The LPIPS metric for the background assesses background preservation, whereas the LPIPS metric for the foreground measures the similarity between the reference texture and the generated content. The value of \textit{Ewarp} falls within the range of 
\(1 \times 10^{-3}\). The best results are \textbf{boldfaced}, and the second-best results are \underline{underlined}.}

\begin{tabular}{
    l|c|c|c|cccc|cccc|c|cc
}
\toprule
\multirow{2}{*}{Method}&\multirow{2}{*}{Stage}&\multirow{2}{*}{Reference}&\multirow{2}{*}{Structure}& \multicolumn{4}{c|}{Background} & \multicolumn{4}{c|}{Foreground} & \multicolumn{1}{c|}{Motion} & \multicolumn{2}{c}{LLM Evaluation} \\ \cline{5-15}
&&&& {MSE$\downarrow$} & {PSNR$\uparrow$} & {SSIM$\uparrow$} & {LPIPS$\downarrow$}
& {CLIP$\uparrow$} & {DINO$\uparrow$} & {LPIPS$\downarrow$} & {Dream$\uparrow$}
& {EWarp $\downarrow$} & {GPT-5$\uparrow$} & {Gemini$\uparrow$} \\ \hline
Ab-1 & \multirow{6}{*}{Stage-1}&w/o Jigsaw & Canny & 68.83& 32.62& 0.9068&0.0800 & 0.7022 & 0.1859 & 0.7674 & 0.7046 & 1.5998 &2.10 & 2.36\\ \cline{3-15}
Ab-2 && \multirow{4}{*}{w/Jigsaw}&Canny & \underline{30.65}& \underline{36.47}& \underline{0.9460}& 0.0379  &0.7149 & 0.1906 & 0.7347 & 0.7205 & 1.4582 & 2.42 & 2.76\\ 
Ab-3 && &HED & 30.69& 36.40& 0.9459& 0.0379&0.6976 & 0.1990 & 0.7484 & 0.7080 & \textbf{1.2395}&2.44&2.74\\
Ab-4 && &Gray   & 30.70& 36.29& 0.9458& 0.0379&0.7182 & 0.2115 & 0.7016 & 0.7352 & 1.4502&2.66&2.94\\
Ab-5 && &Depth  & 30.73& 36.08& 0.9458& \underline{0.0378}&0.6894 & 0.1790 & 0.7532 & 0.7017 & 1.3764&2.21&2.47\\ \cline{3-15}
Ab-6 && w/ Jigsaw&Untextured Video   & 30.66 & 36.44 & \underline{0.9460} & 0.0379
& \underline{0.7331} & \underline{0.2622} & \underline{0.6540} & \underline{0.7473}
& \underline{1.3510} & \underline{2.72}& \textbf{3.27}\\ \hline 
Ab-7 &Stage-2& w/ Jigsaw&Untextured Video   & \textbf{30.35} & \textbf{36.48} & \textbf{0.9485} & \textbf{0.0344} & \textbf{0.7524} & \textbf{0.3241} & \textbf{0.6080} & \textbf{0.7742} & 1.4248& \textbf{2.82}& \underline{3.25}\\ \hline 
\bottomrule
\end{tabular}

\label{tab:ablation-video-table}
\end{table*}

\begin{table*}[htbp]
\centering
\scriptsize
\setlength{\tabcolsep}{6.75pt}
\renewcommand{\arraystretch}{1.18}

\caption{Ablation study for patch size in Jigsaw Permutation. The LPIPS metric for the background assesses background preservation, whereas the LPIPS metric for the foreground measures the similarity between the reference texture and the generated content. The value of \textit{Ewarp} falls within the range of \(1 \times 10^{-3}\). The best results are \textbf{boldfaced}, and the second-best results are \underline{underlined}.}

\begin{tabular}{
    l|c|cccc|cccc|c|cc
}
\toprule
\multirow{2}{*}{Method}&\multirow{2}{*}{Patch Size}&\multicolumn{4}{c|}{Background} & \multicolumn{4}{c|}{Foreground} & \multicolumn{1}{c|}{Motion} & \multicolumn{2}{c}{LLM Evaluation} \\ \cline{3-13}
&& {MSE$\downarrow$} & {PSNR$\uparrow$} & {SSIM$\uparrow$} & {LPIPS$\downarrow$}
& {CLIP$\uparrow$} & {DINO$\uparrow$} & {LPIPS$\downarrow$} & {Dream$\uparrow$}
& {EWarp $\downarrow$} & {GPT-5$\uparrow$} & {Gemini$\uparrow$} \\ \hline
Ab-1&2\%   & 29.87 & 36.67 & \underline{0.9485} & \textbf{0.0326} & 0.7184 & 0.2158  &0.7023 & 0.7210 & 1.5218 & 2.61 & 3.08\\
Ab-2&5\%   & 29.85 & 36.68 & \underline{0.9485} & \textbf{0.0326} & 0.7276 & 0.2495  &\textbf{0.6387} & \textbf{0.7526} & 1.3996 & \underline{2.76} &\textbf{3.22}\\
Ab-3&10\%  & 29.86 & \underline{36.70} & \underline{0.9485} & \textbf{0.0326} & \underline{0.7305} & \textbf{0.2615} & 0.6504 & \underline{0.7410} & 1.4495 & \underline{2.76}&\underline{3.10}\\
Ab-4&20\%  & \underline{29.84} & 36.69 & \underline{0.9485} & \textbf{0.0326} & \textbf{0.7344} & \underline{0.2500} & 0.6554 & 0.7397 & 1.4603&\textbf{2.78}&3.08\\
Ab-5&50\%  & \textbf{29.83} & \underline{36.70} & \underline{0.9485} & \textbf{0.0326} & 0.7232 & 0.2345 & 0.6586 & 0.7316 & \underline{1.4352} &2.60&\underline{3.10} \\
Ab-6&100\% & \textbf{29.83} & \textbf{36.71} & \textbf{0.9486} & \textbf{0.0326} & 0.7247 & 0.2380 & \underline{0.6503} & 0.7357 & \textbf{1.3970}&\textbf{2.78}&\underline{3.10}\\ \hline
\bottomrule
\end{tabular}
\vspace{-2em}
\label{tab:patch_video-table}
\end{table*}

\subsection{Implementation Details of Texture Remover}
We construct a dataset from 72K distinct object meshes extracted from images with clearly identifiable foreground objects. Each mesh is rendered into short paired video sequences as described in Sec.\ref{sec:texture_remover}. Generating approximately eight pairs per object with different augmentation parameters yields 576K paired videos in total—each consisting of a textured source and texture-free target video.
Our model is initialized from VACE and trained for two epochs (18K steps, ~38 hours) on 32 A800 GPUs with a global batch size of 32, constant learning rate of $1\times10^{-5}$, gradient checkpointing, and mixed-precision training. We further apply DMD distillation (learning rate $5\times10^{-6}$, batch size 8, and 300 steps) to produce a fast Texture Remover requiring only three sampling steps at inference.

\subsection{Implementation Details of Refaçade}

\noindent\textbf{Stage 1: Large-Scale Pretraining.}
We pretrain the model for two epochs on a mixture of (i) filtered subset of WebVid-10M containing 1.8M videos, (ii) 900k synthetic videos generated by SelfForcing~\citep{self_forcing_huang2025self}, and (iii) 800k synthetic images produced by Stable Diffusion 3.5 Large~\citep{stabilityai_sd3_5_large_2024}. The network is initialized from VACE and trained on 96 A800 GPUs with a global batch size of 96 and gradient accumulation of 4, corresponding to 18k training steps over 120 hours. We use a constant learning rate of $1\times10^{-5}$, enable gradient checkpointing, and train with mixed precision.

\noindent\textbf{Stage 2: High-quality Finetuning.}
We finetune the model on 180k real videos from Pexels. Finetuning is run for two epochs on 32 A800 GPUs with a global batch size of 32 and gradient accumulation of 4, yielding 2.8k training steps with 28 hours. We keep the same training hyperparameters as in Stage 1, including a constant learning rate of $1\times10^{-5}$, gradient checkpointing, and mixed-precision training.

\subsection{Quantitative Results} \label{sec_quantitative}
We compare  \abbr{} against extensive baselines including specialized inpainting models, general-purpose  editing methods, and closed-source commercial APIs. Results are presented in Tables~\ref{tab:image-table} and~\ref{tab:video-table}. Baseline implementation details are provided in the supplementary materials.

\noindent\textbf{Benchmark Details}
Our evaluation benchmark is organized as quadruples, each consisting of a source image/video, a mask, a reference image, and a prompt.
For image evaluation, we use the high-resolution image dataset UHRSD~\citep{xie2022pyramid}, which contains 988 images and their corresponding masks. We then employ Flux Kontext to generate reference images with salient objects and randomly pair them with the sources. Qwen2.5-VL 32B~\citep{Qwen2.5-VL} takes both the source and the reference image as input to produce captions, from which we derive an instructive prompt and a descriptive prompt that serve as text conditions for some of the methods.
For video evaluation, we use 50 videos from Pexels as the test set, which is disjoint from our training data. The reference images are obtained in the same way as for images, using the first frame of each video for captioning.

\noindent\textbf{Automatic Evaluation}. We evaluate background preservation using MSE, PSNR, SSIM, and LPIPS, and foreground fidelity using CLIPScore~\citep{clipscore_hessel2021clipscore}, DINO~\citep{oquab2023dinov2}, LPIPS~\citep{lpips_zhang2018unreasonable}, and DreamSim~\citep{fu2023dreamsim}. Video motion consistency is assessed via EWarp~\citep{lai2018learning}. As shown in Table~\ref{tab:image-table}, our stage2 model achieves superior background preservation on the image benchmark, substantially outperforming strong baselines such as Flux Fill. Foreground metrics further demonstrate our advantage, with stage2 model attaining the highest CLIPScore (0.7774), DINO (0.4516), and DreamSim (0.8184), alongside the lowest LPIPS (0.6181). On the video benchmark (Table~\ref{tab:video-table}), stage2 model again achieves optimal background reconstruction, surpassing VideoPainter. Foreground alignment improves substantially, while temporal stability remains competitive (EWarp: 1.4248 vs. 1.3510 for stage1). Overall, our framework establishes state-of-the-art performance on both image and video texture transfer through high-fidelity background preservation, semantically consistent foreground editing, and strong temporal coherence.

\noindent\textbf{LLM-based Evaluation}. To address limitations of automatic metrics in capturing perceptual quality, we employ GPT-5 and Gemini-2.5 for evaluation. LLMs are instructed to evaluate the results along four dimensions: (i) whether the generated texture matches that of the reference image;
(ii) whether the generated color is consistent with the reference image;
(iii) whether the object structure in the result remains consistent with the source; and
(iv) whether the background is preserved as in the source image. Our stage2 model consistently ranks highest: on images, it scores 2.89 with GPT-5 and 2.77 with Gemini-2.5, compared with 2.71 and 1.98 for Flux Fill and 2.65 and 2.41 for NanoBanana; on videos, it achieves 2.82 with GPT-5 and 3.25 with Gemini-2.5, versus 2.21 and 2.18 for AnyV2V.

\noindent\textbf{User Study}. To further validate our approach with human judgment, we conduct an extensive user study. We compare the outputs of all competing methods on both images and videos and invite users to evaluate the edited results. Participants are shown the source image/video, the reference image, and the outputs of different methods. They are then asked to evaluate the outputs along three dimensions: (i) whether the reference material is successfully transferred to the selected object; (ii) whether the background is preserved; and (iii) whether the object's structure is maintained. The user preferences on the image and video benchmarks are reported in Tables~\ref{tab:image-table} and~\ref{tab:video-table}, respectively. Our method consistently receives the highest number of votes for both images and videos.

\subsection{Qualitative Results}
Visual comparisons in Figure~\ref{fig:image_video_compare} demonstrate superior background preservation, texture coherence, and foreground fidelity, validating the effectiveness of our framework in terms of perceptual quality. Our method excels in three key aspects:
(i) it edits the entire object, unlike HiDream-E1 and NanoBanana;
(ii) it precisely preserves the background, outperforming Qwen-Image-Edit and InsVIE; and
(iii) it achieves better texture consistency with the reference image during retexturing.

\subsection{Ablation Study}
To validate our design choices, we conduct ablation studies on structural conditioning, jigsaw augmentation, patch size, and two-stage training, as shown in Tables~\ref{tab:ablation-video-table} and~\ref{tab:patch_video-table}.

\noindent\textbf{Impact of Structural Conditions}. 
Table~\ref{tab:ablation-video-table} compares different structural conditioning signals (Ab-2 to Ab-6). Although all variants exhibit comparable background preservation (similar MSE and SSIM), their foreground quality diverges notably. Conventional signals such as Canny, HED, grayscale, and depth produce weaker texture transfer, whereas our untextured video conditioning (Ab-6) consistently achieves higher semantic alignment and lower perceptual distortion, as reflected by CLIP score, LPIPS, DINO score, and LLM-based scores. This indicates that traditional structural cues are prone to texture leakage, where residual appearance information contaminates the conditioning and ultimately degrades texture transfer.

\noindent\textbf{Impact of Jigsaw Permutation}.
Table~\ref{tab:ablation-video-table} compares Ab-1 (without jigsaw) and Ab-2 (with jigsaw). Without jigsaw augmentation, performance degrades across all metrics: background MSE increases from 30.65 to 68.83, PSNR drops from 36.47 to 32.62, and foreground LPIPS worsens from 0.7347 to 0.7674. LLM scores also decline (GPT-5: 2.10 vs. 2.42). This demonstrates that jigsaw augmentation is essential for preventing \textit{geometry leakage}, where the reference image's structure contaminates the output.

\noindent\textbf{Impact of Patch Size in Jigsaw Permutation}.
Table~\ref{tab:patch_video-table} examines the effect of patch size, where the value is expressed as a percentage of the reference image side length (i.e., the side length of each square patch divided by the side length of the full reference frame). In particular, the 100\% setting degenerates to using the original reference image without any jigsaw permutation. Across all settings, background preservation remains similar, while foreground quality varies. Small patches (2\%) yield weaker alignment, medium patches (5–10\%) achieve a better texture transfer, whereas large patches (50–100\%) provide the best temporal stability at the cost of slightly reduced texture fidelity. 

\noindent\textbf{Impact of Two-Stage Training}.
As shown in Table~\ref{tab:ablation-video-table}, stage two Ab-7 improves upon stage one Ab-6 across all metrics. Background LPIPS decreases from 0.0379 to 0.0344, foreground DINO score increases from 0.2622 to 0.3241, and LPIPS improves from 0.6540 to 0.6080. LLM scores under GPT-5 also rise from 2.72 to 2.82, confirming that the second stage effectively refines texture transfer quality.

\section{Conclusion}

In this paper,
we introduce \abbr{} for a new editing task, object retexture.  Our method is designed to enhance controllability and suppress unwanted information during texture transfer. It comprises two key components. First, we replace traditional control conditions with texture-free representations rendered from 3D object meshes, which preserve the structural information of the original object while excluding color and texture cues. Second, we introduce a jigsaw permutation strategy that disrupts spatial structure in the reference image, forcing the model to attend to texture statistics rather than object layout.
Extensive experiments demonstrate that our approach can accurately transfer the target texture onto source objects while preserving their structure, and produces visually compelling results.

\small
\bibliographystyle{ieeenat_fullname}
\bibliography{main}

@String(CVPR  = {IEEE Conf. Comput. Vis. Pattern Recog.})

@String(ICCV  = {Int. Conf. Comput. Vis.})

@String(ECCV  = {Eur. Conf. Comput. Vis.})

@String(NeurIPS = {Adv. Neural Inform. Process. Syst.})

@String(ICML  = {Int. Conf. Mach. Learn.})

@String(ICLR  = {Int. Conf. Learn. Represent.})

@String(AAAI  = {AAAI})

@String(TMLR  = {Trans. Mach. Learn Res.})

@String(CVPR  = {CVPR})

@String(ICCV  = {ICCV})

@String(ECCV  = {ECCV})

@String(NeurIPS = {NeurIPS})

@String(ICML  = {ICML})

@String(ICLR  = {ICLR})

@String(TMLR  = {TMLR})

@String(CVPR= {IEEE Conf. Comput. Vis. Pattern Recog.})

@String(ICCV= {Int. Conf. Comput. Vis.})

@String(ECCV= {Eur. Conf. Comput. Vis.})

@String(ICLR = {Int. Conf. Learn. Represent.})

@String(AAAI = {AAAI})

@inproceedings{lai2018learning,
  title={Learning blind video temporal consistency},
  author={Lai, Wei-Sheng and Huang, Jia-Bin and Wang, Oliver and Shechtman, Eli and Yumer, Ersin and Yang, Ming-Hsuan},
  booktitle={Proceedings of the European conference on computer vision (ECCV)},
  pages={170--185},
  year={2018}
}

@inproceedings{controlnet_zhang2023adding,
  title={Adding conditional control to text-to-image diffusion models},
  author={Zhang, Lvmin and Rao, Anyi and Agrawala, Maneesh},
  booktitle={ICCV},
  year={2023}
}

@inproceedings{groundingdino_liu2023grounding,
  title        = {Grounding DINO: Marrying DINO with Grounded Pre-Training for Open-Set Object Detection},
  author       = {Liu, Shilong and Zeng, Zhaoyang and Ren, Tianhe and Li, Feng and Zhang, Hao and Yang, Jie and Li, Chunyuan and Yang, Jianwei and Su, Hang and Zhu, Jun and Zhang, Lei},
  booktitle    = {ECCV},
  year         = {2024}
}

@inproceedings{animatediff_guo2023animatediff,
  title        = {AnimateDiff: Animate Your Personalized Text-to-Image Diffusion Models without Specific Tuning},
  author       = {Guo, Yuwei and Yang, Ceyuan and Rao, Anyi and Liang, Zhengyang and Wang, Yaohui and Qiao, Yu and Agrawala, Maneesh and Lin, Dahua and Dai, Bo},
  booktitle    = {ICLR},
  year         = {2024}
}

@inproceedings{sdxl_podell2023sdxl,
  title={SDXL: improving latent diffusion models for high-resolution image synthesis},
  author={Podell, Dustin and English, Zion and Lacey, Kyle and Blattmann, Andreas and Dockhorn, Tim and M{\"u}ller, Jonas and Penna, Joe and Rombach, Robin},
  booktitle={ICLR},
  year={2024}
}

@misc{gen3,
  author = {{Gen-3}},
  title = {Introducing Gen-3 Alpha: A New Frontier for Video Generation},
  howpublished = {\url{https://runwayml.com/research/introducing-gen-3-alpha/}},
  year = {2024}
}

@inproceedings{cogvideox_yang2024cogvideox,
  title     = {CogVideoX: Text-to-Video Diffusion Models with An Expert Transformer},
  author    = {Yang, Zhuoyi and Teng, Jiayan and Zheng, Wendi and Ding, Ming and Huang, Shiyu and Xu, Jiazheng and Yang, Yuanming and Hong, Wenyi and Zhang, Xiaohan and Feng, Guanyu and Yin, Da and Gu, Xiaotao and Zhang, Yuxuan and Wang, Weihan and Cheng, Yean and Liu, Ting and Xu, Bin and Dong, Yuxiao and Tang, Jie},
  booktitle = {ICLR},
  year      = {2025}
}

@inproceedings{brushnet_ju2024brushnetplugandplayimageinpainting,
  title     = {BrushNet: A Plug-and-Play Image Inpainting Model with Decomposed Dual-Branch Diffusion},
  author    = {Ju, Xuan and Liu, Xian and Wang, Xintao and Bian, Yuxuan and Shan, Ying and Xu, Qiang},
  booktitle = {ECCV},
  year      = {2024}
}

@inproceedings{sam2_ravi2025sam2,
  title     = {SAM 2: Segment Anything in Images and Videos},
  author    = {Ravi, Nikhila and Gabeur, Valentin and Hu, Yuan-Ting and Hu, Ronghang and Ryali, Chaitanya and Ma, Tengyu and Khedr, Haitham and Rädle, Roman and Rolland, Chloe and Gustafson, Laura and Mintun, Eric and Pan, Junting and Alwala, Kalyan Vasudev and Carion, Nicolas and Wu, Chao-Yuan and Girshick, Ross and Dollár, Piotr and Feichtenhofer, Christoph},
  booktitle = {ICLR},
  year      = {2025}
}

@misc{pexels,
  author="Pexels",
  title="https://www.pexels.com/",
  url = "https://www.pexels.com/",
  year = "2024"
}

@inproceedings{dit_peebles2022scalable,
  title     = {Scalable Diffusion Models with Transformers},
  author    = {Peebles, William and Xie, Saining},
  booktitle = {ICCV},
  year      = {2023}
}

@misc{mochi_1,
  author = {{Mochi-1}},
  title = {Mochi-1},
  howpublished = {\url{https://www.genmo.ai/blog}},
  year = {2024}
}

@inproceedings{webvid_bain2021frozen,
  title     = {Frozen in Time: A Joint Video and Image Encoder for End-to-End Retrieval},
  author    = {Bain, Max and Nagrani, Arsha and Varol, G{\"u}l and Zisserman, Andrew},
  booktitle = {ICCV},
  year      = {2021}
}

@article{avid_zhang2023avid,
  title   = {AVID: Any-Length Video Inpainting with Diffusion Model},
  author  = {Zhang, Zhixing and Wu, Bichen and Wang, Xiaoyan and Luo, Yaqiao and Zhang, Luxin and Zhao, Yinan and Vajda, Peter and Metaxas, Dimitris and Yu, Licheng},
  journal = {arXiv:2312.03816},
  year    = {2023}
}

@inproceedings{imagen_saharia2022photorealistic,
  title     = {Photorealistic Text-to-Image Diffusion Models with Deep Language Understanding},
  author    = {Saharia, Chitwan and Chan, William and Saxena, Saurabh and Li, Lala and Whang, Jay and Denton, Emily L and Ghasemipour, Kamyar and Gontijo Lopes, Raphael and Karagol Ayan, Burcu and Salimans, Tim and others},
  booktitle = {NeurIPS},
  year      = {2022}
}

@inproceedings{tuneavideo_wu2023tune,
  title     = {Tune-A-Video: One-Shot Tuning of Image Diffusion Models for Text-to-Video Generation},
  author    = {Wu, Jay Zhangjie and Ge, Yixiao and Wang, Xintao and Lei, Stan Weixian and Gu, Yuchao and Shi, Yufei and Hsu, Wynne and Shan, Ying and Qie, Xiaohu and Shou, Mike Zheng},
  booktitle = {ICCV},
  year      = {2023}
}

@inproceedings{stablediffusion_blattmann2023align,
  title     = {Align Your Latents: High-Resolution Video Synthesis with Latent Diffusion Models},
  author    = {Blattmann, Andreas and Rombach, Robin and Ling, Huan and Dockhorn, Tim and Kim, Seung Wook and Fidler, Sanja and Kreis, Karsten},
  booktitle = {CVPR},
  year      = {2023}
}

@inproceedings{smartbrush_xie2023smartbrush,
  title={Smartbrush: Text and shape guided object inpainting with diffusion model},
  author={Xie, Shaoan and Zhang, Zhifei and Lin, Zhe and Hinz, Tobias and Zhang, Kun},
  booktitle={CVPR},
  year={2023}
}

@inproceedings{clipscore_hessel2021clipscore,
  author = {Hessel, Jack and Holtzman, Ari and Forbes, Maxwell and Le Bras, Ronan and Choi, Yejin},
  title = {CLIPScore: A Reference-free Evaluation Metric for Image Captioning},
  booktitle = {EMNLP},
  year = {2021}
}

@inproceedings{tokenflow_geyer2023tokenflow,
  author = {Geyer, Michal and Bar-Tal, Omer and Bagon, Shai and Dekel, Tali},
  title = {TokenFlow: Consistent Diffusion Features for Consistent Video Editing},
  booktitle = {ICLR},
  year = {2024}
}

@inproceedings{imageneditor_wang2023imagen,
  title={Imagen editor and editbench: Advancing and evaluating text-guided image inpainting},
  author={Wang, Su and Saharia, Chitwan and Montgomery, Ceslee and Pont-Tuset, Jordi and Noy, Shai and Pellegrini, Stefano and Onoe, Yasumasa and Laszlo, Sarah and Fleet, David J and Soricut, Radu and others},
  booktitle={CVPR},
  year={2023}
}

@article{ku2024anyv2v,
  author = {Ku, Max and Wei, Cong and Ren, Weiming and Yang, Huan and Chen, Wenhu},
  title = {AnyV2V: A plug-and-play framework for any video-to-video editing tasks},
  journal = {TMLR},
  year = {2024}
}

@misc{sora,
  author = {OpenAI},
  title = {Sora: Creating video from text},
  howpublished = {\url{https://openai.com/index/sora/}},
  year = {2024}
}

@inproceedings{chen2024videocrafter2,
  title={Videocrafter2: Overcoming data limitations for high-quality video diffusion models},
  author={Chen, Haoxin and Zhang, Yong and Cun, Xiaodong and Xia, Menghan and Wang, Xintao and Weng, Chao and Shan, Ying},
  booktitle={CVPR},
  year={2024}
}

@inproceedings{cong2023flatten,
  author = {Cong, Yuren and Xu, Mengmeng and Simon, Christian and Chen, Shoufa and Ren, Jiawei and Xie, Yanping and Perez-Rua, Juan-Manuel and Rosenhahn, Bodo and Xiang, Tao and He, Sen},
  title = {FLATTEN: optical flow-guided attention for consistent text-to-video editing},
  booktitle = {ICLR},
  year = {2024}
}

@article{wang2024videocomposer,
  title={Videocomposer: Compositional video synthesis with motion controllability},
  author={Wang, Xiang and Yuan, Hangjie and Zhang, Shiwei and Chen, Dayou and Wang, Jiuniu and Zhang, Yingya and Shen, Yujun and Zhao, Deli and Zhou, Jingren},
  journal={NeurIPS},
  year={2024}
}

@inproceedings{pixart_chen2023,
  author = {Chen, Junsong and Yu, Jincheng and Ge, Chongjian and Yao, Lewei and Xie, Enze and Wu, Yue and Wang, Zhongdao and Kwok, James and Luo, Ping and Lu, Huchuan and others},
  title = {PixArt-α: Fast Training of Diffusion Transformer for Photorealistic Text-to-Image Synthesis},
  booktitle = {ICLR},
  year = {2024}
}

@misc{flux,
  author = {{Black Forest Labs}},
  title = {Black Forest Labs},
  howpublished = {\url{https://github.com/black-forest-labs/flux/}},
  year = {2024}
}

@inproceedings{omni_citation_geng2024instructdiffusion,
  title={Instructdiffusion: A generalist modeling interface for vision tasks},
  author={Geng, Zigang and Yang, Binxin and Hang, Tiankai and Li, Chen and Gu, Shuyang and Zhang, Ting and Bao, Jianmin and Zhang, Zheng and Li, Houqiang and Hu, Han and others},
  booktitle={CVPR},
  year={2024}
}

@inproceedings{instructpix2pix_brooks2023instructpix2pix,
  title={Instructpix2pix: Learning to follow image editing instructions},
  author={Brooks, Tim and Holynski, Aleksander and Efros, Alexei A},
  booktitle={CVPR},
  year={2023}
}

@article{magic_brush_zhang2024magicbrush,
  title={Magicbrush: A manually annotated dataset for instruction-guided image editing},
  author={Zhang, Kai and Mo, Lingbo and Chen, Wenhu and Sun, Huan and Su, Yu},
  journal={NeurIPS},
  year={2024}
}

@inproceedings{ultraedit_zhao2024ultraedit,
  author = {Zhao, Haozhe and Ma, Xiaojian and Chen, Liang and Si, Shuzheng and Wu, Rujie and An, Kaikai and Yu, Peiyu and Zhang, Minjia and Li, Qing and Chang, Baobao},
  title = {UltraEdit: Instruction-based Fine-Grained Image Editing at Scale},
  booktitle = {NeurIPS},
  year = {2024}
}

@inproceedings{video_p2p_liu2024video,
  title={Video-p2p: Video editing with cross-attention control},
  author={Liu, Shaoteng and Zhang, Yuechen and Li, Wenbo and Lin, Zhe and Jia, Jiaya},
  booktitle={CVPR},
  year={2024}
}

@inproceedings{
insv2v_cheng2024consistent,
title={Consistent Video-to-Video Transfer Using Synthetic Dataset},
author={Jiaxin Cheng and Tianjun Xiao and Tong He},
booktitle={ICLR},
year={2024}
}

@inproceedings{omniedit_wei2024omniedit,
  author = {Wei, Cong and Xiong, Zheyang and Ren, Weiming and Du, Xinrun and Zhang, Ge and Chen, Wenhu},
  title = {OmniEdit: Building Image Editing Generalist Models Through Specialist Supervision},
  booktitle = {ICLR},
  year = {2025}
}

@inproceedings{hq_edit_hui2024hq,
  author = {Hui, Mude and Yang, Siwei and Zhao, Bingchen and Shi, Yichun and Wang, Heng and Wang, Peng and Zhou, Yuyin and Xie, Cihang},
  title = {HQ-Edit: A High-Quality Dataset for Instruction-based Image Editing},
  booktitle = {ICLR},
  year = {2025}
}

@inproceedings{cococo_zi2024cococo,
  author = {Zi, Bojia and Zhao, Shihao and Qi, Xianbiao and Wang, Jianan and Shi, Yukai and Chen, Qianyu and Liang, Bin and Wong, Kam-Fai and Zhang, Lei},
  title = {CoCoCo: Improving Text-Guided Video Inpainting for Better Consistency, Controllability and Compatibility},
  booktitle = {AAAI},
  year = {2025}
}

@article{vivid_10m_hu2024vivid,
  title={VIVID-10M: A Dataset and Baseline for Versatile and Interactive Video Local Editing},
  author={Hu, Jiahao and Zhong, Tianxiong and Wang, Xuebo and Jiang, Boyuan and Tian, Xingye and Yang, Fei and Wan, Pengfei and Zhang, Di},
  journal={arXiv:2411.15260},
  year={2024}
}

@article{Cogvlm2_hong2024cogvlm2,
  title={Cogvlm2: Visual language models for image and video understanding},
  author={Hong, Wenyi and Wang, Weihan and Ding, Ming and Yu, Wenmeng and Lv, Qingsong and Wang, Yan and Cheng, Yean and Huang, Shiyu and Ji, Junhui and Xue, Zhao and others},
  journal={arXiv:2408.16500},
  year={2024}
}

@article{hunyuanvideo_kong2024hunyuanvideo,
  title={Hunyuanvideo: A systematic framework for large video generative models},
  author={Kong, Weijie and Tian, Qi and Zhang, Zijian and Min, Rox and Dai, Zuozhuo and Zhou, Jin and Xiong, Jiangfeng and Li, Xin and Wu, Bo and Zhang, Jianwei and others},
  journal={arXiv:2412.03603},
  year={2024}
}

@article{wan2025,
      title={Wan: Open and Advanced Large-Scale Video Generative Models}, 
      author={Ang Wang and Baole Ai and Bin Wen and Chaojie Mao and Chen-Wei Xie and Di Chen and Feiwu Yu and Haiming Zhao and Jianxiao Yang and Jianyuan Zeng and Jiayu Wang and Jingfeng Zhang and Jingren Zhou and Jinkai Wang and Jixuan Chen and Kai Zhu and Kang Zhao and Keyu Yan and Lianghua Huang and Mengyang Feng and Ningyi Zhang and Pandeng Li and Pingyu Wu and Ruihang Chu and Ruili Feng and Shiwei Zhang and Siyang Sun and Tao Fang and Tianxing Wang and Tianyi Gui and Tingyu Weng and Tong Shen and Wei Lin and Wei Wang and Wei Wang and Wenmeng Zhou and Wente Wang and Wenting Shen and Wenyuan Yu and Xianzhong Shi and Xiaoming Huang and Xin Xu and Yan Kou and Yangyu Lv and Yifei Li and Yijing Liu and Yiming Wang and Yingya Zhang and Yitong Huang and Yong Li and You Wu and Yu Liu and Yulin Pan and Yun Zheng and Yuntao Hong and Yupeng Shi and Yutong Feng and Zeyinzi Jiang and Zhen Han and Zhi-Fan Wu and Ziyu Liu},
      journal = {arXiv:2503.20314},
      year={2025}
}

@inproceedings{senorita_zi2025se,
  title={Senorita-2M: A High-Quality Instruction-based Dataset for General Video Editing by Video Specialists},
  author={Zi, Bojia and Ruan, Penghui and Chen, Marco and Qi, Xianbiao and Hao, Shaozhe and Zhao, Shihao and Huang, Youze and Liang, Bin and Xiao, Rong and Wong, Kam-Fai},
  booktitle={NeurIPS Dataset and Benchmark Track},
  year={2025}
}

@inproceedings{minimax_remover_zi2025minimax,
  title={MiniMax-Remover: Taming Bad Noise Helps Video Object Removal},
  author={Zi, Bojia and Peng, Weixuan and Qi, Xianbiao and Wang, Jianan and Zhao, Shihao and Xiao, Rong and Wong, Kam-Fai},
  booktitle={NeurIPS},
  year={2025}
}

@inproceedings{ff_vdi_lee2025_ff_vdi_video,
  title={Video diffusion models are strong video inpainter},
  author={Lee, Minhyeok and Cho, Suhwan and Shin, Chajin and Lee, Jungho and Yang, Sunghun and Lee, Sangyoun},
  booktitle={AAAI},
  year={2025}
}

@article{floed_gu2024advanced,
  title={Advanced Video Inpainting Using Optical Flow-Guided Efficient Diffusion},
  author={Gu, Bohai and Luo, Hao and Guo, Song and Dong, Peiran},
  journal={arXiv:2412.00857},
  year={2024}
}

@article{diffueraser_li2025diffueraser,
  title={DiffuEraser: A Diffusion Model for Video Inpainting},
  author={Li, Xiaowen and Xue, Haolan and Ren, Peiran and Bo, Liefeng},
  journal={arXiv:2501.10018},
  year={2025}
}

@inproceedings{bian2025videopainter,
  author = {Bian, Yuxuan and Zhang, Zhaoyang and Ju, Xuan and Cao, Mingdeng and Xie, Liangbin and Shan, Ying and Xu, Qiang},
  title = {VideoPainter: Any-length Video Inpainting and Editing with Plug-and-Play Context Control},
  booktitle = {SIGGRAPH},
  year = {2025}
}

@article{vace_jiang2025vace,
  title={Vace: All-in-one video creation and editing},
  author={Jiang, Zeyinzi and Han, Zhen and Mao, Chaojie and Zhang, Jingfeng and Pan, Yulin and Liu, Yu},
  journal={arXiv:2503.07598},
  year={2025}
}

@article{mtv_inpaint_yang2025mtv,
  title={MTV-Inpaint: Multi-Task Long Video Inpainting},
  author={Yang, Shiyuan and Gu, Zheng and Hou, Liang and Tao, Xin and Wan, Pengfei and Chen, Xiaodong and Liao, Jing},
  journal={arXiv:2503.11412},
  year={2025}
}

@article{stepvideo_ma2025step,
  title={Step-video-t2v technical report: The practice, challenges, and future of video foundation model},
  author={Ma, Guoqing and Huang, Haoyang and Yan, Kun and Chen, Liangyu and Duan, Nan and Yin, Shengming and Wan, Changyi and Ming, Ranchen and Song, Xiaoniu and Chen, Xing and others},
  journal={arXiv:2502.10248},
  year={2025}
}

@article{step1x_edit_liu2025step1x_edit,
      title={Step1X-Edit: A Practical Framework for General Image Editing}, 
      author={Shiyu Liu and Yucheng Han and Peng Xing and Fukun Yin and Rui Wang and Wei Cheng and Jiaqi Liao and Yingming Wang and Honghao Fu and Chunrui Han and Guopeng Li and Yuang Peng and Quan Sun and Jingwei Wu and Yan Cai and Zheng Ge and Ranchen Ming and Lei Xia and Xianfang Zeng and Yibo Zhu and Binxing Jiao and Xiangyu Zhang and Gang Yu and Daxin Jiang},
      journal={arXiv:2504.17761},
      year={2025}
}

@inproceedings{wu2025insvie,
  title={InsViE-1M: Effective Instruction-based Video Editing with Elaborate Dataset Construction},
  author={Wu, Yuhui and Chen, Liyi and Li, Ruibin and Wang, Shihao and Xie, Chenxi and Zhang, Lei},
  booktitle={ICCV},
  year={2025}
}

@inproceedings{flow_match_lipman2023flow,
  title={Flow Matching for Generative Modeling},
  author={Lipman, Yotam and Ho, Jonathan and Salimans, Tim and Carmon, Yair and Duvenaud, David and Isola, Phillip and Doron, Gabi},
  booktitle={ICML},
  year={2023}
}

@inproceedings{powerpaint_zhuang2024task,
  title={A task is worth one word: Learning with task prompts for high-quality versatile image inpainting},
  author={Zhuang, Junhao and Zeng, Yanhong and Liu, Wenran and Yuan, Chun and Chen, Kai},
  booktitle={ECCV},
  year={2024}
}

@misc{qwenimage_wu2025qwenimagetechnicalreport,
      title={Qwen-Image Technical Report}, 
      author={Chenfei Wu and Jiahao Li and Jingren Zhou and Junyang Lin and Kaiyuan Gao and Kun Yan and Sheng-ming Yin and Shuai Bai and Xiao Xu and Yilei Chen and Yuxiang Chen and Zecheng Tang and Zekai Zhang and Zhengyi Wang and An Yang and Bowen Yu and Chen Cheng and Dayiheng Liu and Deqing Li and Hang Zhang and Hao Meng and Hu Wei and Jingyuan Ni and Kai Chen and Kuan Cao and Liang Peng and Lin Qu and Minggang Wu and Peng Wang and Shuting Yu and Tingkun Wen and Wensen Feng and Xiaoxiao Xu and Yi Wang and Yichang Zhang and Yongqiang Zhu and Yujia Wu and Yuxuan Cai and Zenan Liu},
      year={2025},
      eprint={arXiv:2508.02324}
}

@inproceedings{dmd2_yin2024dmd2,
  title={Improved Distribution Matching Distillation for Fast Image Synthesis},
  author={Yin, Tianwei and Gharbi, Micha{\"e}l and Park, Taesung and Zhang, Richard and Shechtman, Eli and Durand, Fr{\'e}do and Freeman, William T.},
  booktitle={NeurIPS},
  year={2024}
}

@misc{vipaint_agarwal2024vipaintimageinpaintingpretrained,
      title={VIPaint: Image Inpainting with Pre-Trained Diffusion Models via Variational Inference}, 
      author={Sakshi Agarwal and Gabe Hoope and Erik B. Sudderth},
      year={2024},
      eprint={arXiv:2411.18929}
}

@misc{ControlNetInpaint,
  author       = {mikonvergence},
  title        = {ControlNetInpaint: Inpaint images with ControlNet},
  howpublished = {\url{https://github.com/mikonvergence/ControlNetInpaint}},
  year         = {2023}
}

@inproceedings{turbofill_xie2025turbofill,
  title={TurboFill: Adapting Few-step Text-to-image Model for Fast Image Inpainting},
  author={Xie, Liangbin and Pakhomov, Daniil and Wang, Zhonghao and Wu, Zongze and Chen, Ziyan and Zhou, Yuqian and Zheng, Haitian and Zhang, Zhifei and Lin, Zhe and Zhou, Jiantao and others},
  booktitle={CVPR},
  year={2025}
}

@inproceedings{attentive_eraser_sun2025attentive,
  title={Attentive eraser: Unleashing diffusion model’s object removal potential via self-attention redirection guidance},
  author={Sun, Wenhao and Dong, Xue-Mei and Cui, Benlei and Tang, Jingqun},
  booktitle={AAAI},
  year={2025}
}

@inproceedings{designedit_Jia_Cheng_Yuan_Wang_Li_Jia_Zhang_2025, title={DesignEdit: Unify Spatial-Aware Image Editing via Training-free Inpainting with a Multi-Layered Latent Diffusion Framework}, booktitle={AAAI}, author={Jia, Yueru and Cheng, Aosong and Yuan, Yuhui and Wang, Chuke and Li, Ji and Jia, Huizhu and Zhang, Shanghang}, year={2025}
}

@inproceedings{rorem_li2025rorem,
  title={RORem: Training a Robust Object Remover with Human-in-the-Loop},
  author={Li, Ruibin and Yang, Tao and Guo, Song and Zhang, Lei},
  booktitle={CVPR},
  year={2025}
}

@article{objectclear_zhao2025objectclear,
  title={ObjectClear: Complete Object Removal via Object-Effect Attention},
  author={Zhao, Jixin and Zhou, Shangchen and Wang, Zhouxia and Yang, Peiqing and Loy, Chen Change},
  journal={arXiv:2505.22636},
  year={2025}
}

@article{zhao2025hunyuan3d,
  title={Hunyuan3d 2.0: Scaling diffusion models for high resolution textured 3d assets generation},
  author={Zhao, Zibo and Lai, Zeqiang and Lin, Qingxiang and Zhao, Yunfei and Liu, Haolin and Yang, Shuhui and Feng, Yifei and Yang, Mingxin and Zhang, Sheng and Yang, Xianghui and others},
  journal={arXiv:2501.12202},
  year={2025}
}

@article{cai2025hidream,
  title={HiDream-I1: A High-Efficient Image Generative Foundation Model with Sparse Diffusion Transformer},
  author={Cai, Qi and Chen, Jingwen and Chen, Yang and Li, Yehao and Long, Fuchen and Pan, Yingwei and Qiu, Zhaofan and Zhang, Yiheng and Gao, Fengbin and Xu, Peihan and others},
  journal={arXiv:2505.22705},
  year={2025}
}

@article{kolors,
  title={Kolors: Effective Training of Diffusion Model for Photorealistic Text-to-Image Synthesis},
  author={Kolors Team},
  journal={preprint},
  year={2024}
}

@article{self_forcing_huang2025self,
  title={Self Forcing: Bridging the Train-Test Gap in Autoregressive Video Diffusion},
  author={Huang, Xun and Li, Zhengqi and He, Guande and Zhou, Mingyuan and Shechtman, Eli},
  journal={arXiv:2506.08009},
  year={2025}
}

@misc{stabilityai_sd3_5_large_2024,
  title        = {Introducing Stable Diffusion 3.5},
  author       = {{Stability AI Team}},
  year         = {2024},
  howpublished = {\url{https://stability.ai/news/introducing-stable-diffusion-3-5}},
  note         = {Accessed 2025-10-28}
}

@inproceedings{mmdit_esser2024scaling,
  title={Scaling rectified flow transformers for high-resolution image synthesis},
  author={Esser, Patrick and Kulal, Sumith and Blattmann, Andreas and Entezari, Rahim and M{\"u}ller, Jonas and Saini, Harry and Levi, Yam and Lorenz, Dominik and Sauer, Axel and Boesel, Frederic and others},
  booktitle={ICML},
  year={2024}
}

@inproceedings{sd3_esser2024scaling,
  title={Scaling rectified flow transformers for high-resolution image synthesis},
  author={Esser, Patrick and Kulal, Sumith and Blattmann, Andreas and Entezari, Rahim and M{\"u}ller, Jonas and Saini, Harry and Levi, Yam and Lorenz, Dominik and Sauer, Axel and Boesel, Frederic and others},
  booktitle={ICML},
  year={2024}
}

@misc{google_nanobanana,
  title        = {Nano Banana - Gemini AI image generator \& photo editor},
  author       = {Google DeepMind},
  year         = {2025},
  howpublished = {\url{https://gemini.google/overview/image-generation/}}
}

@misc{blackforestlabs2025fluxkontext,
  title        = {FLUX.1 Kontext: State-of-the-art in-context image generation and editing},
  author       = {Black Forest Labs},
  year         = {2025},
  howpublished = {\url{https://bfl.ai/models/flux-kontext}}
}

@article{hidreami1technicalreport,
  title={HiDream-I1: A High-Efficient Image Generative Foundation Model with Sparse Diffusion Transformer},
  author={Cai, Qi and Chen, Jingwen and Chen, Yang and Li, Yehao and Long, Fuchen and Pan, Yingwei and Qiu, Zhaofan and Zhang, Yiheng and Gao, Fengbin and Xu, Peihan and others},
  journal={arXiv:2505.22705},
  year={2025}
}

@article{ditto_bai2025scaling,
  title={Scaling Instruction-Based Video Editing with a High-Quality Synthetic Dataset},
  author={Bai, Qingyan and Wang, Qiuyu and Ouyang, Hao and Yu, Yue and Wang, Hanlin and Wang, Wen and Cheng, Ka Leong and Ma, Shuailei and Zeng, Yanhong and Liu, Zichen and others},
  journal={arXiv preprint arXiv:2510.15742},
  year={2025}
}

@article{icve_liao2025context,
  title={In-Context Learning with Unpaired Clips for Instruction-based Video Editing},
  author={Liao, Xinyao and Zeng, Xianfang and Song, Ziye and Fu, Zhoujie and Yu, Gang and Lin, Guosheng},
  journal={arXiv:2510.14648},
  year={2025}
}

@misc{decartai2025lucyedit,
  title        = {Lucy Edit: Open-weight Text-guided Video Editing},
  author       = {DecartAI Team},
  year         = {2025},
  note         = {Accessed: 2025-11-13}
}

@article{ju2025editverse,
  title={EditVerse: Unifying Image and Video Editing and Generation with In-Context Learning},
  author={Ju, Xuan and Wang, Tianyu and Zhou, Yuqian and Zhang, He and Liu, Qing and Zhao, Nanxuan and Zhang, Zhifei and Li, Yijun and Cai, Yuanhao and Liu, Shaoteng and others},
  journal={arXiv:2509.20360},
  year={2025}
}

@article{oquab2023dinov2,
  title={Dinov2: Learning robust visual features without supervision},
  author={Oquab, Maxime and Darcet, Timoth{\'e}e and Moutakanni, Th{\'e}o and Vo, Huy and Szafraniec, Marc and Khalidov, Vasil and Fernandez, Pierre and Haziza, Daniel and Massa, Francisco and El-Nouby, Alaaeldin and others},
  journal={arXiv:2304.07193},
  year={2023}
}

@inproceedings{fu2023dreamsim,
  title     = {DreamSim: Learning New Dimensions of Human Visual Similarity using Synthetic Data},
  author    = {Stephanie Fu and Netanel Tamir and Shobhita Sundaram and Lucy Chai and Richard Zhang and Tali Dekel and Phillip Isola},
  booktitle = {NeurIPS},
  year      = {2023}
}

@inproceedings{lpips_zhang2018unreasonable,
  title={The unreasonable effectiveness of deep features as a perceptual metric},
  author={Zhang, Richard and Isola, Phillip and Efros, Alexei A and Shechtman, Eli and Wang, Oliver},
  booktitle={CVPR},
  year={2018}
}

@inproceedings{xie2022pyramid,
    author    = {Xie, Chenxi and Xia, Changqun and Ma, Mingcan and Zhao, Zhirui and Chen, Xiaowu and Li, Jia},
    title     = {Pyramid Grafting Network for One-Stage High Resolution Saliency Detection},
    booktitle = {CVPR},
    year      = {2022}
}

@article{Qwen2.5-VL,
  title={Qwen2.5-VL Technical Report},
  author={Bai, Shuai and Chen, Keqin and Liu, Xuejing and Wang, Jialin and Ge, Wenbin and Song, Sibo and Dang, Kai and Wang, Peng and Wang, Shijie and Tang, Jun and Zhong, Humen and Zhu, Yuanzhi and Yang, Mingkun and Li, Zhaohai and Wan, Jianqiang and Wang, Pengfei and Ding, Wei and Fu, Zheren and Xu, Yiheng and Ye, Jiabo and Zhang, Xi and Xie, Tianbao and Cheng, Zesen and Zhang, Hang and Yang, Zhibo and Xu, Haiyang and Lin, Junyang},
  journal={arXiv:2502.13923},
  year={2025}
}

@article{zhao2023unipc,
  title={UniPC: A Unified Predictor-Corrector Framework for Fast Sampling of Diffusion Models},
  author={Zhao, Wenliang and Bai, Lujia and Rao, Yongming and Zhou, Jie and Lu, Jiwen},
  journal={NeurIPS},
  year={2023}
}

@article{shi2015hierarchical,
  title={Hierarchical image saliency detection on extended CSSD},
  author={Shi, Jianping and Yan, Qiong and Xu, Li and Jia, Jiaya},
  journal={TPAMI},
  year={2015}
}

@article{qwen3technicalreport,
      title={Qwen3 Technical Report}, 
      author={Qwen Team},
      year={2025},
      journal={arXiv:2505.09388}
}

@inproceedings{Perazzi_CVPR_2016,
  author    = {Federico Perazzi and
               Jordi Pont-Tuset and
               Brian McWilliams and
               Luc Van Gool and
               Markus Gross and
               Alexander Sorkine-Hornung},
  title     = {A Benchmark Dataset and Evaluation Methodology for Video Object Segmentation},
  booktitle = {CVPR},
  year      = {2016}
}

@article{2023i2vgenxl,
  title={I2VGen-XL: High-Quality Image-to-Video Synthesis via Cascaded Diffusion Models},
  author={Zhang, Shiwei and Wang, Jiayu and Zhang, Yingya and Zhao, Kang and Yuan, Hangjie and Qing, Zhiwu and Wang, Xiang  and Zhao, Deli and Zhou, Jingren},
  journal={arXiv:2311.04145},
  year={2023}
}

@article{comanici2025gemini2_5,
  title={Gemini 2.5: Pushing the frontier with advanced reasoning, multimodality, long context, and next generation agentic capabilities},
  author={Comanici, Gheorghe and Bieber, Eric and Schaekermann, Mike and Pasupat, Ice and Sachdeva, Noveen and Dhillon, Inderjit and Blistein, Marcel and Ram, Ori and Zhang, Dan and Rosen, Evan and others},
  journal={arXiv:2507.06261},
  year={2025}
}

@misc{openai_gpt5_2025,
  author       = {OpenAI},
  title        = {{GPT-5 is here}},
  year         = {2025}
}

\clearpage
\appendix

\section{Related Works}
\subsection{Image Inpainting}
Recent progress in image editing is largely driven by image diffusion models~\citep{stablediffusion_blattmann2023align, flux, sdxl_podell2023sdxl, brushnet_ju2024brushnetplugandplayimageinpainting, vipaint_agarwal2024vipaintimageinpaintingpretrained}. Among these methods, SD-Inpainting~\citep{stablediffusion_blattmann2023align} and ControlNet Inpainting~\citep{ControlNetInpaint} extend Stable Diffusion by fine-tuning on datasets of randomly masked images paired with text prompts. Although these adaptations generate visually plausible results, they often drift from the input text and struggle to place objects accurately according to the described semantics. To mitigate this issue, SmartBrush~\citep{smartbrush_xie2023smartbrush} and Imagen Editor~\citep{imageneditor_wang2023imagen} incorporate paired object-description data, yet they implicitly assume that the masked region always contains an object, which restricts their capacity for context-aware completion. PowerPaint~\citep{powerpaint_zhuang2024task} instead learns task-specific prompts that adapt to the mask, which strengthens the relationship between textual input and contextual surroundings and leads to state-of-the-art performance in both context-aware inpainting and text-guided editing. BrushNet~\citep{brushnet_ju2024brushnetplugandplayimageinpainting} builds on ControlNet to extract conditioning information and inject it into a frozen diffusion U-Net, whereas Turbo-Fill~\citep{turbofill_xie2025turbofill} emphasizes efficiency by combining a few-step text-to-image diffusion process with an inpainting adapter to achieve fast and high-fidelity results. Flux-Fill~\citep{flux}, trained on the Flux base model, likewise produces visually compelling inpainting outcomes. In addition, several methods focus on editing flexibility and object removal. Attentive Eraser~\citep{attentive_eraser_sun2025attentive} proposes a tuning-free strategy that enables pre-trained diffusion models to perform stable and effective object removal. DesignEdit~\citep{designedit_Jia_Cheng_Yuan_Wang_Li_Jia_Zhang_2025} introduces a simple yet powerful approach for spatially flexible editing that first inpaints the background and then applies a two-stage multi-layer latent diffusion framework to modify each element independently. RORem~\citep{rorem_li2025rorem} adopts a semi-supervised human-in-the-loop pipeline to curate high-quality paired training data, and ObjectClear~\citep{objectclear_zhao2025objectclear} integrates an object-effect attention mechanism that guides the model toward target foreground regions through learned attention masks.
\subsection{Video Inpainting}
Analogous to image inpainting, existing video inpainting approaches can be broadly grouped into video object removal and text-guided video inpainting. Within video object removal, a line of work focuses on explicit removal of target objects. FFF-VDI~\citep{ff_vdi_lee2025_ff_vdi_video} propagates future-frame latents to initialize masked regions and then fine-tunes an image-to-video diffusion model to complete the corrupted area. FloED~\citep{floed_gu2024advanced} injects both optical-flow and text embeddings to guide removal. DiffuEraser~\citep{diffueraser_li2025diffueraser} couples flow-guided inpainting with DDIM inversion to attain higher fidelity. Senorita-Remover~\citep{senorita_zi2025se} relies on instruction-driven prompts, using positive prompts to guide removal and negative prompts to suppress unintended content. Minimax-Remover~\citep{minimax_remover_zi2025minimax} employs a minimax optimization objective that improves removal quality and prevents undesired object regeneration. For text-guided video inpainting, recent work addresses masked-region generation and editing under text prompts. VideoComposer~\citep{wang2024videocomposer} is an early diffusion model for text-guided video inpainting that offers multi-conditional control within a unified framework. AVID~\citep{avid_zhang2023avid} scales to sequences of arbitrary length from natural-language prompts. COCOCO~\citep{cococo_zi2024cococo} improves consistency and controllability using damped global attention and stronger text cross-attention. VIVID~\citep{vivid_10m_hu2024vivid} provides a 10M-scale image video corpus for localized editing, which enables more capable text-guided inpainters. MTV-Inpaint~\citep{mtv_inpaint_yang2025mtv} unifies scene completion and novel object insertion within a single framework. VideoPainter~\citep{bian2025videopainter} adopts a DiT-based architecture with a context encoder that injects background cues into a pretrained video DiT to achieve plug-and-play consistent inpainting. More recently, VACE~\citep{vace_jiang2025vace} introduces a video editing framework that consumes multiple control signals to generate edited videos.

\noindent \textbf{Remark.} \emph{Despite notable successes, most inpainting systems remain unable to use a reference image to direct the outcome inside the missing areas.}

\section{Dataset Setup Details}

The dataset used to train our Stage 1 model consists of two components, a filtered subset of WebVid-10M and our synthetic dataset. The former provides large-scale and inexpensive video resources, while the latter focuses on videos containing objects with rare and long-tailed textures. To construct the synthetic dataset, we use Qwen3-14B~\citep{qwen3technicalreport} to generate 2.2M prompts, which are then used for text-to-image synthesis with Stable Diffusion 3.5 Large~\citep{stabilityai_sd3_5_large_2024} and text-to-video synthesis with Self-Forcing~\citep{self_forcing_huang2025self}. The Stage 2 model is trained on videos from Pexels\citep{pexels}, which leads to improved aesthetic quality.

\noindent \textbf{Dataset for ablation study}. For the patch size ablation in Table 4 of the main text, we consider a setting where the reference image shares the same texture as the target object but differs in shape and size. We denote this setting as a patch size of 100\%. To obtain such paired images, we employ Flux-Kontext~\citep{blackforestlabs2025fluxkontext} and leverage its image-to-image capability to generate reference images by reshaping the target objects. We use the following prompt:

\begin{tcolorbox}[colback=white,colframe=black!75!white,
  title=Prompt for Reference Image Generation,fonttitle=\bfseries]
\small
Reshape the \texttt{\{source object\}} into a \texttt{\{target object\}} with same color and texture, white background.
\end{tcolorbox}

\begin{figure}
    \centering
    \includegraphics[width=0.985\linewidth]{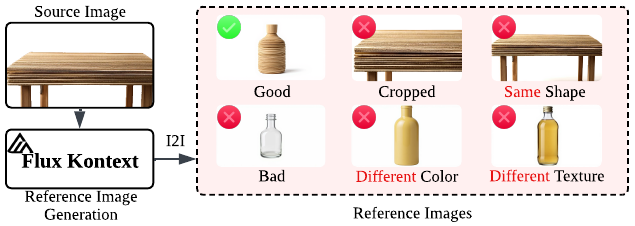}
    \caption{Visual results of reference images generated by Flux-Kontext for our ablation study. The results can be divided into six categories, and only those reference images that have a different shape from the source but share the same material and dominant color are retained.
}
    \label{fig:flux-kontext}
\end{figure}

However, relying solely on Flux-Kontext to generate reference images is not sufficient, as we observed that many of the resulting references are suboptimal as shown in Figure~\ref{fig:flux-kontext}. Therefore, we further filter the generated images and videos using GPT-5~\citep{openai_gpt5_2025}, which yields a final dataset of only 10K video pairs and 8K image pairs for no-Jigsaw training.

\section{Implementation Details}

\subsection{Training Details of Refaçade and Texture Remover}
\textbf{Training Details of Refaçade}
During training, we randomly resize and downsample frames. In addition, we randomly drop the conditioning information with probability 0.1 by replacing the reference image with an all-white image and its mask with all-black pixel value, so that classifier-free guidance can be applied at inference time. The batch size is 96 in Stage 1 and 32 in Stage 2, with constant learning rate of 1e-5. 

\noindent \textbf{Training Details of Texture Remover}
The training procedure for the Texture Remover is similar to that of Refaçade. We use DMD2 to distill the Texture Remover from 50 sampling steps to 3 steps.
\noindent Table~\ref{tab:refacade_detail} summarizes the key hyperparameters of Refaçade and the Texture Remover.

\begin{table}
\caption{Hyperparameter of \abbr{} and Texture Remover. }
\scriptsize
\label{tab:refacade_detail}
\centering
\begin{tabular}{l|cccc}
\toprule
\multirow{3}{*}{\raisebox{-2.0ex}{Config}} & \multicolumn{4}{c}{\textbf{Model}} \\
\cmidrule(lr){2-5}
 & \multicolumn{2}{c}{\abbr{}} & \multicolumn{2}{c}{\textbf{Texture Remover}} \\
\cmidrule(lr){2-3}\cmidrule(lr){4-5}
 & Stage 1 & Stage 2 & Stage 1 & Distill \\
\midrule
Batch Size / GPU  & \multicolumn{2}{c}{1} & \multicolumn{2}{c}{1}\\
Accumulation Step  & \multicolumn{2}{c}{4} & \multicolumn{2}{c}{1}\\
Gradient Checkpointing & \multicolumn{2}{c}{True} & \multicolumn{2}{c}{True}\\
Optimizer  & \multicolumn{2}{c}{AdamW} & \multicolumn{2}{c}{AdamW} \\
Learning Rate  & \multicolumn{2}{c}{$1\times10^{-5}$} &$1\times10^{-5}$ &$5\times10^{-6}$ \\
LR Schedule  & \multicolumn{2}{c}{Constant} & \multicolumn{2}{c}{Constant}\\
Time Sampling  & \multicolumn{2}{c}{Uniform} & \multicolumn{2}{c}{Uniform}\\
Num GPUs & 96 & 32 & 32 & 8 \\
Training Steps &18000 &2800 & 18000 & 300\\
\midrule
Num Main Layers & \multicolumn{2}{c}{24} & \multicolumn{2}{c}{24}\\
Token Dimension  & \multicolumn{2}{c}{1536} & \multicolumn{2}{c}{1536}\\
Parameters  & \multicolumn{2}{c}{2.0258B} & \multicolumn{2}{c}{1.7143B}\\
Control Layer Indices & \multicolumn{2}{c}{0,5,10,15,20,24,28} & \multicolumn{2}{c}{0,5,10,15,20,25} \\
Pre-trained Model & \multicolumn{2}{c}{Wan2.1-VACE-1.3B} & \multicolumn{2}{c}{Wan2.1-VACE-1.3B}\\
\midrule
Sample Steps & \multicolumn{2}{c}{20} & 50 &3 \\
Sampler & \multicolumn{2}{c}{Flow UniPC~\citep{zhao2023unipc}} & \multicolumn{2}{c}{Flow Euler}\\
\textbf{Input Resolution(s)} 
  & \multicolumn{2}{c}{Multi-resolution} 
  & \multicolumn{2}{c}{Multi-resolution} \\
\textbf{Frame Count(s)} 
  & \multicolumn{2}{c}{Multiple frame lengths} 
  & \multicolumn{2}{c}{Multiple frame lengths} \\
\bottomrule
\end{tabular}
\end{table}

\subsection{Inference Details of Texture Remover}
At inference time, we provide the object mask together with the input video and remove the background so that the input matches the training format. The Texture Remover then produces a sequence of texture free mesh videos that are temporally aligned with the source video. The same pipeline also supports single frame inference, where a single reference image is treated as a video of length one, which allows us to obtain texture free three dimensional control conditions from still images. During inference, we disable classifier free guidance and use three sampling steps. For video inputs, inference is performed at a resolution of 480$\times$832 with 81 frames, and for image inputs it is performed at the original resolution.

\subsection{Inference Details of Refaçade}
Inference is conducted on a single RTX 4090.

\noindent \textbf{Image editing.}
Editing a single image with resolution $480\times 832$ peaks at about 12 GB of GPU memory and takes approximately $6.5$ s. The Texture Remover accounts for about $0.35$ s, and the remaining overhead arises from VAE encoding and decoding as well as the diffusion process. The CFG scale is set to $1.5$.

\noindent \textbf{Video editing.}
Editing an 81 frame video at $480\times 832$ peaks at about 20 GB of GPU memory and takes approximately $150$ s. The Texture Remover contributes $5.5$ s of this runtime. The CFG scale is set to $1.5$.

\subsection{Inference Details of Baseline Methods}

Most image and video editors rely on text prompts rather than reference images. To accommodate such editors, we use Qwen-VL-2.5 32B to generate captions for the reference image and then convert these captions into text prompts. For editors that accept a reference image as input, we directly feed the reference image into the model for inference. The templates for generating prompts are detailed below.

\begin{tcolorbox}[colback=white,colframe=black!75!white,
  title=Template for Instructive Prompt Generation,fonttitle=\bfseries,breakable]
\small
You are given an image and a target object name: \texttt{\{object\_name\}}.

1) Identify the dominant color tone(s) and the surface material/texture of the main object in the image (choose the largest/central salient object).

2) Write ONE imperative instruction to restyle \texttt{\{object\_name\}} with that exact color and material.

Rules:
- 20--40 words, one sentence, NO ENTER.

- Mention color shade (e.g., dark brown, icy blue) and material (e.g., chocolate texture, brushed metal, glossy ceramic).

- No extra commentary.

For example, ``Turn the dog into dark brown, covered with chocolate texture''. 
If multiple objects, ``Turn the bike and man into dark brown, covered with chocolate texture''.

Please ONLY return the instructive prompt sentence.
\end{tcolorbox}

\begin{tcolorbox}[colback=white,colframe=black!75!white,
  title=Template for Descriptive Prompt Generation,fonttitle=\bfseries,breakable]
\small
You are given an image and a target object name: \texttt{\{object\_name\}}.

1) Identify the dominant color tone(s) and the surface material/texture of the main object in the image (choose the largest/central salient object).

2) Write ONE descriptive prompt to describe \texttt{\{object\_name\}} with that exact color and material.
Rules:

- 20–40 words, one sentence, NO ENTER.

- Mention color shade (e.g., dark brown, icy blue) and material (e.g., chocolate texture, brushed metal, glossy ceramic).

- No extra commentary.

For example, "A dog in dark brown, covered with chocolate texture". If multiple objects, "Bike and man in dark brown, covered with chocolate texture."

Please ONLY return the descriptive prompt sentence.
\end{tcolorbox}

\subsubsection{Image Baseline}

\noindent\textbf{Implementation Details of BrushNet.}
We adopt BrushNet with its released pretrained checkpoint together with the Stable Diffusion XL base model to generate images at resolution $1024 \times 1024$.
The pipeline takes the original image, its corresponding mask and a descriptive prompt as input.
We perform 50 denoising steps with a CFG scale equal to 5.0 and set \texttt{brushnet\_conditioning\_scale} to 1.0.
The generated images are then resized to the original resolution for comparison with other methods.

\noindent\textbf{Implementation Details of Controlnet-Inpainting.}
We use the pretrained control block checkpoint together with the Stable Diffusion 1.5 base model for inference.
Input images are first resized to resolution $512 \times 512$.
We then provide the source image and its corresponding mask together with a descriptive prompt to the inference pipeline.
We employ the default settings with 20 denoising steps and a CFG scale is set to 7.5.

\noindent\textbf{Implementation Details of Flux-Fill.}
We use the FLUX.1-Fill-dev model and perform inference at the original image resolution.
The pipeline takes the source image, its corresponding mask and a descriptive prompt as input.
We set the CFG scale to 30.0 and use 50 steps for inference.

\noindent\textbf{Implementation Details of Flux-Kontext-Text.}
We use the FLUX.1-Kontext-dev model conditioned on the instructive prompt.
Inference is performed at the original image resolution with 28 denoising steps and CFG is set to 3.0.

\noindent\textbf{Implementation Details of Flux-Kontext-Image.}
We use the FLUX.1-Kontext-dev model conditioned on both the reference image and the instructive prompt.
Inference is performed at the original resolution of the source image and mask using 42 denoising steps with a CFG scale equal to 2.5 and \texttt{strength} set to 1.0.

\noindent\textbf{Implementation Details of HiDream-E1.}
We use the HiDream-E1-1 model.
The source image and mask are first resized to resolution $768 \times 768$.
We set the CFG to 3, \texttt{image\_guidance\_scale} to 1.0 and \texttt{refine\_strength} to 0.3.
Both the instructive prompt and the descriptive prompt are used as textual conditions as shown below
\begin{tcolorbox}[colback=white,colframe=black!75!white,title=Prompt for HiDream-E1,fonttitle=\bfseries]
\small
Editing Instruction \{\texttt{instructive\_prompt}\} Target Image Description \{\texttt{descriptive\_prompt}\}
\end{tcolorbox}

\noindent\textbf{Implementation Details of HQ-Edit.}
We use the released pretrained checkpoint of HQ-Edit.
Input images are resized to resolution $512 \times 512$ before inference. We set the CFG to 7.0, perform 30 denoising steps and set \texttt{image\_guidance\_scale} to 1.5 while conditioning on the instructive prompt.
Finally, the generated images are resized back to the original resolution for comparison.

\noindent\textbf{Implementation Details of InsP2P.}
We use the released pretrained InsP2P checkpoint for inference. Input images are resized to resolution $512 \times 512$ in advance. We set the text guidance scale \texttt{text\_cfg\_scale} to 7.5 and the image guidance scale \texttt{image\_cfg\_scale} to 1.5 while conditioning the model on the descriptive prompt.
We perform 100 denoising steps.

\noindent\textbf{Implementation Details of NanoBanana.}
We call the official NanoBanana API to generate edited images. Due to the aspect ratio constraint in this API, we first resize input images to resolution $1024 \times 1024$. The output image is then resized back to the original resolution. The model is conditioned on the source image, the reference image, and the following textual prompt:

\begin{tcolorbox}[colback=white,colframe=black!75!white,title=Prompt for NanoBanana,fonttitle=\bfseries]
\small
Keep the background unchanged. Replace the material/texture of the \{\texttt{object}\} in the first image using the material
from the second image (the reference). Output only the edited image. The output size must exactly match the first image.
\end{tcolorbox}

\noindent\textbf{Implementation Details of Qwen-Image-Edit.}
For Qwen-Image-Edit, we perform inference at the original resolution of each input image. 
We run 50 denoising steps with \texttt{true\_cfg\_scale} set to 4.0, conditioning the model on the instruction prompt.

\noindent\textbf{Implementation Details of Stable Diffusion3-Inpainting.}
For Stable Diffusion3-Inpainting, we use the Stable Diffusion3-medium base model. 
Inference is carried out at the original resolution of the source image and its mask, 
using 50 denoising steps with the CFG scale set to 7.0, conditioned on the descriptive prompt.

\noindent\textbf{Implementation Details of UltraEdit.}
For UltraEdit, we use the pretrained UltraEdit checkpoint for inference. 
The source images and masks are uniformly resized to a resolution of \(512 \times 512\) before sampling. 
We run 50 denoising steps with the CFG scale set to 7.5 and \texttt{image\_guidance\_scale} set to 1.5, 
conditioning the model on the descriptive prompt.

\subsubsection{Video Baseline}

\noindent\textbf{Implementation Details of AnyV2V.}
We adopt a two-stage pipeline built upon I2VGen-XL~\citep{2023i2vgenxl}. In the first stage, we apply DDIM inversion with 500 steps to obtain noisy latents from the input video. In the second stage, we use Flux-Fill to edit the first frame, conditioning the generation on both the inverted latents from the first stage and the descriptive prompt. We set \texttt{pnp\_f\_t} = 1, \texttt{pnp\_spatial\_attn\_t} = 1, and \texttt{pnp\_temp\_attn\_t} = 1. The input videos are resized to a spatial resolution of \(512 \times 512\), and the number of frames is truncated to 36. We use a CFG scale of 9.0 and perform 50 denoising steps. Finally, the generated videos are resized back to the original resolution for comparison.

\noindent\textbf{Implementation Details of COCOCO.}
We use the pretrained COCOCO checkpoint together with the Stable Diffusion Inpainting model for inference. The input videos and their masks are resized to a spatial resolution of \(512 \times 512\) and truncated to 33 frames. We set CFG to 10.0 and perform 50 denoising steps, conditioning the model on the descriptive prompt and using a negative prompt of ``worst quality, low quality''. Finally, the generated videos are resized back to the original resolution for comparison.

\noindent\textbf{Implementation Details of Ditto.}
We use a pretrained LoRA on top of the Wan2.1-VACE-14B base model. Inference is performed at a resolution of \(480 \times 832\) with 33 frames, conditioned on the instructive prompt, while keeping all other settings at their default configuration. The generated videos are resized back to the original resolution.

\noindent\textbf{Implementation Details of Flatten.}
We use Stable Diffusion 2.1 as the base model. The input videos are resized to a spatial resolution of \(512 \times 512\) and truncated to 33 frames. We perform 50 denoising steps with CFG set to 15.0 and set \texttt{inject\_step} to 40, conditioning the model on the descriptive prompt. All other settings follow the default configuration.

\noindent\textbf{Implementation Details of ICVE.}
We use the pretrained ICVE checkpoint together with the HunyuanVideo base model. We follow the default parameter configuration, resizing input videos to a resolution of \(240 \times 384\) and truncating them to 33 frames. Inference is performed with 50 denoising steps and CFG set to 6.0, with \texttt{embedded\_cfg\_scale} set to 1.0, conditioning the model on the instructive prompt.

\noindent\textbf{Implementation Details of InsV2V.}
We use the pretrained InsV2V checkpoint for evaluation. Input videos are resized to a resolution of \(384 \times 384\) and truncated to 33 frames. We set \texttt{text\_cfg} to 7.5 and \texttt{img\_cfg} to 1.2, while keeping all other parameters at their default settings. The generated videos are resized back to the original resolution.

\noindent\textbf{Implementation Details of InsVIE.}
We use the pretrained InsVIE checkpoint together with the CogVideoX-2B base model. Input videos are resized to a resolution of \(480 \times 720\) and truncated to 49 frames. The model is conditioned on the instructive prompt, with a negative prompt of ``bad quality'', while all other parameters follow the default configuration.

\noindent\textbf{Implementation Details of LucyEdit.}
We use the Lucy-Edit-1.1-Dev model for evaluation. Input videos are resized to a resolution of \(480 \times 832\) and truncated to 33 frames. We set CFG to 5.0 and condition the model on the instructive prompt, using empty prompt as the negative prompt. All other settings follow the default configuration. Finally, the generated videos are resized back to the original resolution.

\noindent\textbf{Implementation Details of Se\~norita.}
We use the pretrained Se\~norita checkpoint with the CogVideoX-5b-I2V base model for evaluation. Input videos are resized to \(448 \times 768\) and truncated to 33 frames. The first frame is edited by Flux-Fill and then used as the starting frame for generation. We set CFG to 4.0 and perform 50 denoising steps, conditioning the model on the instructive prompt, while keeping all other parameters at their default settings.

\noindent\textbf{Implementation Details of TokenFlow.}
We use Stable Diffusion 2.1 as the base model. In the first stage, we apply DDIM inversion with 50 steps to obtain noisy latents from the input video. In the second stage, we set \texttt{pnp\_f\_t} = 0.8 and \texttt{pnp\_attn\_t} = 0.5. Inference is performed at the original video resolution, with the number of frames truncated to 40. We set CFG to 7.5 and condition the model on the descriptive prompt, using the latents from the first stage to guide 50 denoising steps.

\noindent\textbf{Implementation Details of VACE.}
We use the Wan2.1-VACE-1.3B model for inference. Input videos are resized to a resolution of \(480 \times 832\) and truncated to 33 frames. We set CFG to 3.0, \texttt{context\_scale} to 1.0, and \texttt{shift\_scale} to 1.0, and perform 20 denoising steps, conditioning the model on the descriptive prompt and the reference image. Empirically, we observe that directly feeding the original video causes the model to copy the foreground and ignore the control signals. To mitigate this, we convert the foreground into a scribble-style representation while preserving the original background as input. Finally, the generated videos are resized back to the original resolution.

\noindent\textbf{Implementation Details of VideoPainter.}
We use the pretrained VideoPainter checkpoint with the CogVideoX-5b-I2V base model. Input is resized to a resolution of \(480 \times 720\) and truncated to 49 frames. The first frame is edited using Flux-Fill and serves as the starting frame for generation. During inference, the foreground mask is dilated by 10 pixels. We perform 50 denoising steps with CFG set to 6.0, conditioning the model on the descriptive prompt.

\subsection{Evaluation Metrics Implementation}
\noindent\textbf{Implementation Details of Background Evaluation.} We first dilate original mask by 16 pixels to accommodate the settings of some editing models. We then compute the average MSE, PSNR, SSIM, and LPIPS over the remaining background region. For videos, these metrics are computed on a per-frame basis and then averaged over all frames of all videos.

\noindent\textbf{Implementation Details of Foreground Evaluation.} As discussed in Sec. 3 of the main text, we use CLIPScore, DINO, LPIPS, and DreamSim for foreground evaluation. Specifically, we first crop the foreground regions from the images or videos and resize them to match the spatial resolution of the reference image. For CLIPScore, DINO, and LPIPS, we use their corresponding base models to extract features from both the generated images or videos and the reference image, and then compute the cosine similarity between the two feature vectors, where a larger value indicates higher similarity in material and color.

\noindent\textbf{Implementation Details of LLM Evaluation.}
Given the source image or videos, the reference image, and the output of one method, we ask GPT-5~\citep{openai_gpt5_2025} and Gemini-2.5-Pro~\citep{comanici2025gemini2_5} to assign a score. The instruction is as follows:

\begin{tcolorbox}[colback=white,colframe=black!75!white,title=Template for LLM Evaluation,fonttitle=\bfseries,breakable]
\small
You will receive three images:

A: the original image with a visible outline over the foreground region (for localization only);

B: the reference image that shows the desired material/texture and color;

C: the candidate (edited) image to be evaluated.

Check ONLY the outlined foreground and return one integer 0..4 (number of satisfied criteria):

1) Material application is reasonable and complete.

2) Color is similar to reference.

3) Structure preserved.

4) Background stays the same as the original.

Return ONLY the integer.
\end{tcolorbox}

\begin{figure}[t]
    \centering

    \begin{subfigure}{\linewidth}
        \centering
        \includegraphics[width=0.985\linewidth]{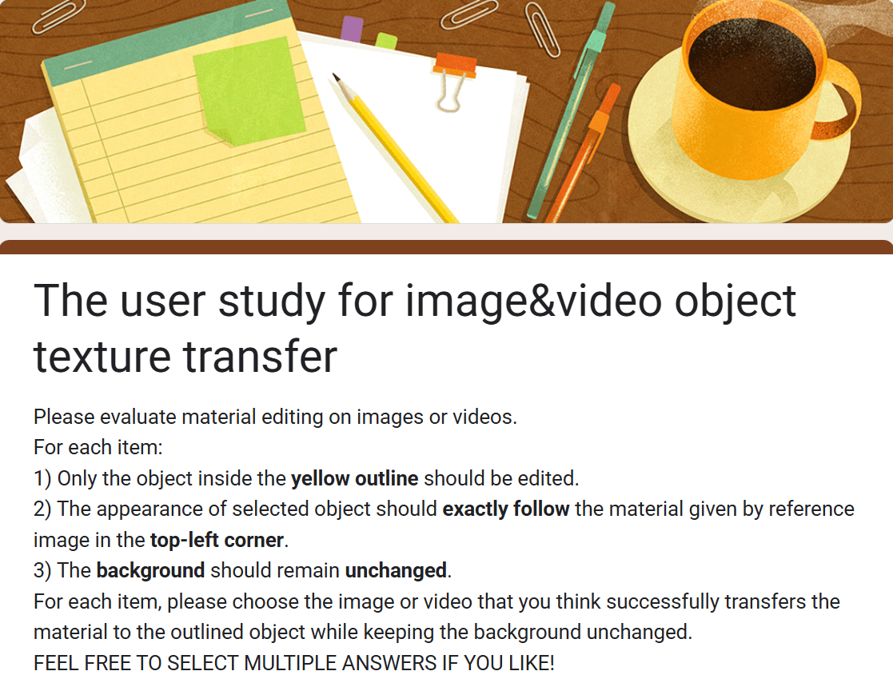}
    \end{subfigure}

    \vspace{-0.5em}

    \begin{subfigure}{\linewidth}
        \centering
        \includegraphics[width=0.985\linewidth]{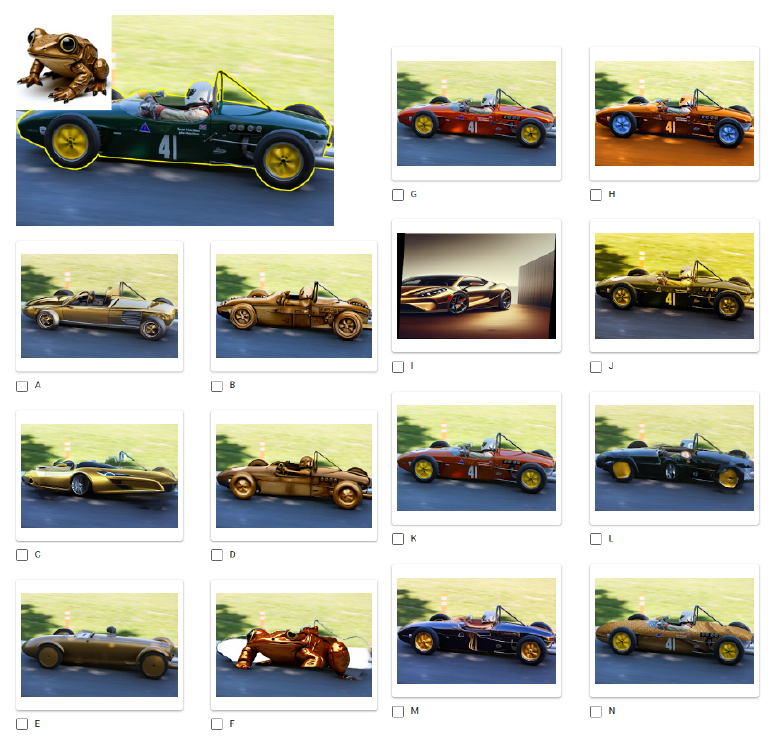}
    \end{subfigure}

    \caption{Questionnaire for user study.}
    \label{fig:questionnaire}
\end{figure}

\noindent\textbf{Implementation Details of User Preference.}
To evaluate human preferences over different editing methods, we design a questionnaire that presents the
results of various image and video editing approaches. Participants are asked to assess the outputs from
multiple aspects and select all options they find satisfactory. The questionnaire instructions are as in Figure~\ref{fig:questionnaire}.

\section{LLM Consistency with Human Annotation}

Following~\citep{minimax_remover_zi2025minimax,omniedit_wei2024omniedit}, we compare the discrepancy between LLM-based scores (GPT-5 and Gemini-2.5 Pro) and human preferences on 90 samples. As shown in Table~\ref{tab:llm_consistency}, Gemini-2.5 Pro~\citep{comanici2025gemini2_5} exhibits preferences that are highly consistent with human annotations, indicating that its scoring criteria are closer to human judgments. GPT-5~\citep{openai_gpt5_2025}, on the other hand, shows larger discrepancies, suggesting that its scores deviate more from human preferences and are generally stricter. Nevertheless, the relative ranking induced by GPT-5 still aligns well with the comparative quality of the different methods.

\begin{table}[h]
  \centering
  \scriptsize
  \caption{LLM Consistency with human annotations.}
  \label{tab:llm_consistency}
  \begin{tabular}{l|ccc}
    \toprule
    Metric & Human & GPT-5 & Gemini-2.5\\
    \midrule
    Score & 3.15  & 2.76 & 3.01 \\
    \bottomrule
  \end{tabular}
\end{table}

\begin{figure}[t]
    \centering
    \begin{subfigure}{\linewidth}
        \centering
        {%
          \fontsize{8pt}{9.6pt}\selectfont
          \begin{tabular}{@{}*{4}{>{\centering\arraybackslash}m{0.20\linewidth}}@{}}
            \minibox{Source} & \minibox{Reference} & \minibox{scale=1.0} & \minibox{scale=1.5}\\
          \end{tabular}
        }
        \vspace{2pt}
        \includegraphics[width=\linewidth]{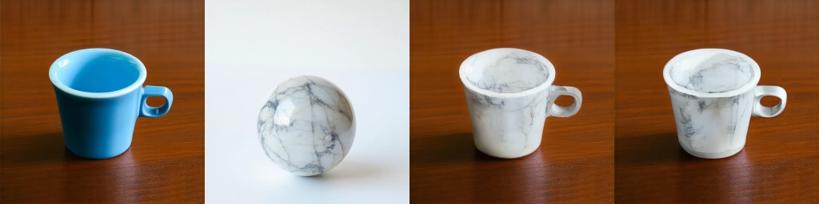}
    \end{subfigure}

    \begin{subfigure}{\linewidth}
        \centering
        {%
          \fontsize{8pt}{9.6pt}\selectfont
          \begin{tabular}{@{}*{4}{>{\centering\arraybackslash}m{0.20\linewidth}}@{}}
            \minibox{scale=2.0} & \minibox{scale=2.5} & \minibox{scale=3.0} & \minibox{scale=4.0}\\
          \end{tabular}
        }
        \vspace{2pt}
        \includegraphics[width=\linewidth]{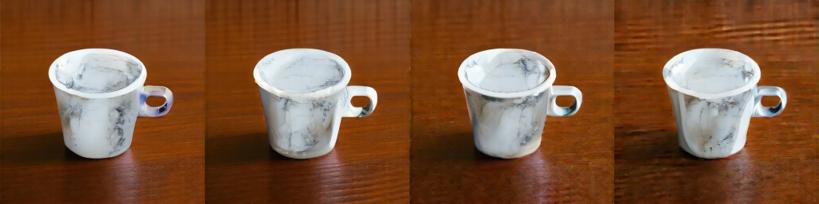}
    \end{subfigure}

    \caption{Qualitative visualization of Refaçade with CFG scales. }
    \label{fig:cfg_comparision}
\end{figure}

\begin{table*}[htbp]
\caption{Ablation study for CFG scales. The LPIPS for background evaluates background perseveration, while LPIPS for foreground evaluates the similarity between reference texture and generated content. CLIP, DINO and Dream are the abbreviations of CLIPScore, DINOScore and DreamSim, respectively. The best results are \textbf{boldfaced}, and the second-best results are \underline{underlined}.}
\scriptsize
\label{tab:cfg_all}
\centering
\setlength{\tabcolsep}{6.5pt}
\renewcommand{\arraystretch}{1.25}
\begin{tabular}{c|cccc|cccc|cc}
\toprule
\multirow{2}{*}{Scale} 
& \multicolumn{4}{c|}{Background} 
& \multicolumn{4}{c|}{Foreground} 
& \multicolumn{2}{c}{LLM Evaluation} \\
\cline{2-11}
& MSE$\downarrow$ & PSNR$\uparrow$ & SSIM$\uparrow$ & LPIPS$\downarrow$
& CLIP$\downarrow$ & DINO$\uparrow$ & LPIPS$\downarrow$ & Dream$\uparrow$
& GPT-5$\uparrow$ & Gemini-2.5$\uparrow$ \\
\toprule
1.0  
& \textbf{28.68} & \textbf{36.82} & \textbf{0.9500} & \textbf{0.0322}
& 0.7224 & 0.2386 & 0.6627 & 0.7303
& 2.640 & 3.080 \\
1.5  
& \underline{29.87} & \underline{36.69} & \underline{0.9485} & \underline{0.0326}
& \underline{0.7331} & \underline{0.2622} & \underline{0.6540} & \underline{0.7473}
& \textbf{2.763} & \textbf{3.280} \\
2.0  
& 30.82 & 36.61 & 0.9470 & 0.0333
& 0.7296 & \textbf{0.2653} & \textbf{0.6526} & 0.7403
& \underline{2.680} & \underline{3.260} \\
2.5  
& 37.41 & 35.46 & 0.9402 & 0.0429
& \textbf{0.7364} & 0.2535 & 0.6680 & \textbf{0.7488}
& 2.760 & 3.020 \\
3.0  
& 101.04 & 30.85 & 0.8970 & 0.0916
& 0.7323 & 0.2559 & 0.6707 & 0.7387
& 2.580 & 2.980 \\
\bottomrule
\end{tabular}
\end{table*}

\section{The Impact of Classifer Free Guidance for Refaçade}

To evaluate the impact of the CFG scale, we conduct experiments on the Pexels validation set. In particular, a scale of 1 corresponds to using only the conditional information without any unconditional guidance. As shown in Table~\ref{tab:cfg_all}, increasing the CFG scale leads to a degradation in background quality. When the CFG scale lies between 1.0 and 2.0, the background remains relatively stable, but it deteriorates rapidly once the scale exceeds 2.5. For the foreground region, we observe that the best scores are mostly concentrated around scales of 1.5 and 2.0, where the model achieves strong material and color similarity to the reference. When the scale exceeds 2.5, the gains in material similarity become marginal and can even turn negative. Figure~\ref{fig:cfg_comparision} illustrates editing results on the same image under different CFG scales. When the scale is set to 1.0, the influence of the reference image is relatively weak: the marble streaks on the cup are sparse, and large regions remain white. When the scale reaches 1.5 or higher, the marble texture becomes much more pronounced. However, when CFG $\geq$ 2.5, background distortions begin to appear; for example, the wooden texture of the table becomes noticeably darker. At a scale of 4.0, clear artifacts can be observed.

\section{Performance of the Texture Remover}

To evaluate the performance of the Texture Remover, we render 50 pairs of textured and texture-free videos as our evaluation dataset, each at a resolution of $480\times832$ with 33 frames. We use exactly the same camera parameters and object motion for each pair to ensure strict correspondence.

\noindent \textbf{Performance of the original Texture Remover.}
As shown in Table~\ref{tab:original_tr}, for the original Texture Remover, increasing the number of inference steps reduces the reconstruction error, but at the cost of substantially higher computation time. In practice, using 50 inference steps is impractical, which highlights the importance of distilling the model to operate reliably with fewer steps.

\begin{figure}[t]
    \centering
    \begin{subfigure}{\linewidth}
        \centering
        {%
          \fontsize{8pt}{9.6pt}\selectfont
          \begin{tabular}{@{}*{3}{>{\centering\arraybackslash}m{0.28\linewidth}}@{}}
            \minibox{Source} & \minibox{Ground Truth} & \minibox{scale=1.0} \\
          \end{tabular}
        }
        \vspace{2pt}
        \includegraphics[width=\linewidth]{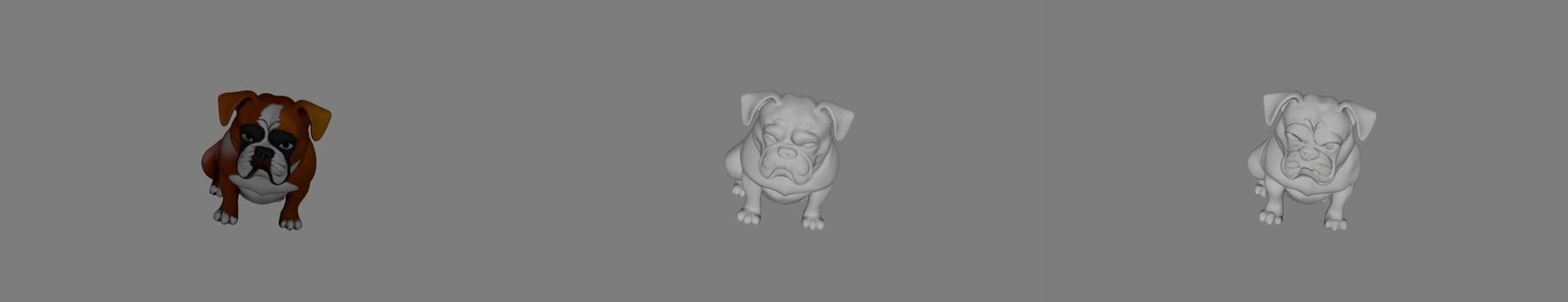}
    \end{subfigure}

    \begin{subfigure}{\linewidth}
        \centering
        {%
          \fontsize{8pt}{9.6pt}\selectfont
          \begin{tabular}{@{}*{3}{>{\centering\arraybackslash}m{0.28\linewidth}}@{}}
            \minibox{scale=1.5} & \minibox{scale=2.0} & \minibox{scale=2.5} \\
          \end{tabular}
        }
        \vspace{2pt}
        \includegraphics[width=\linewidth]{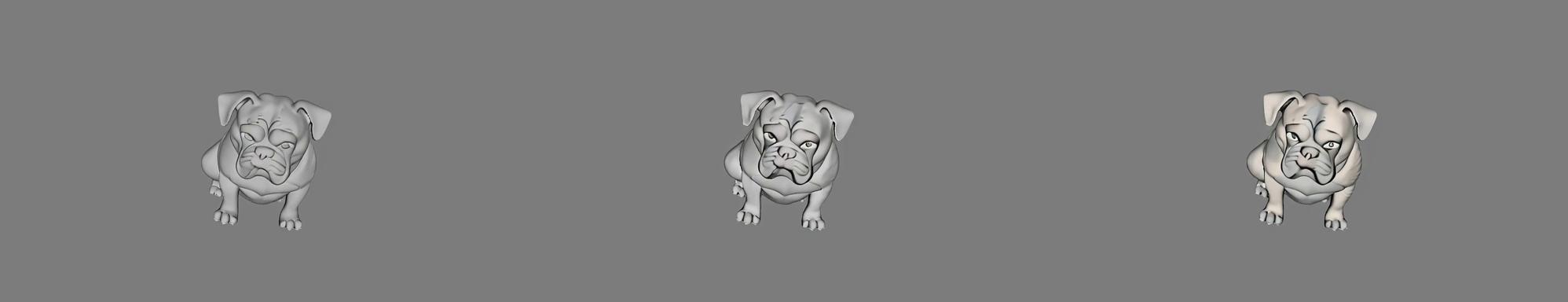}
    \end{subfigure}

    \caption{Qualitative visualization of texture remover with different CFG scales. }
    \label{fig:tr_cfg}
\end{figure}

\begin{figure}[t]
    \centering
    {%
      \fontsize{8pt}{9.6pt}\selectfont
      \begin{tabular}{@{}*{3}{>{\centering\arraybackslash}m{0.135\textwidth}}@{}}
        \minibox{Ground Truth} & 
        \minibox{Primitive (3 steps)} & 
        \minibox{Distilled (3 steps)} \\
      \end{tabular}
    }

    \includegraphics[width=\linewidth]{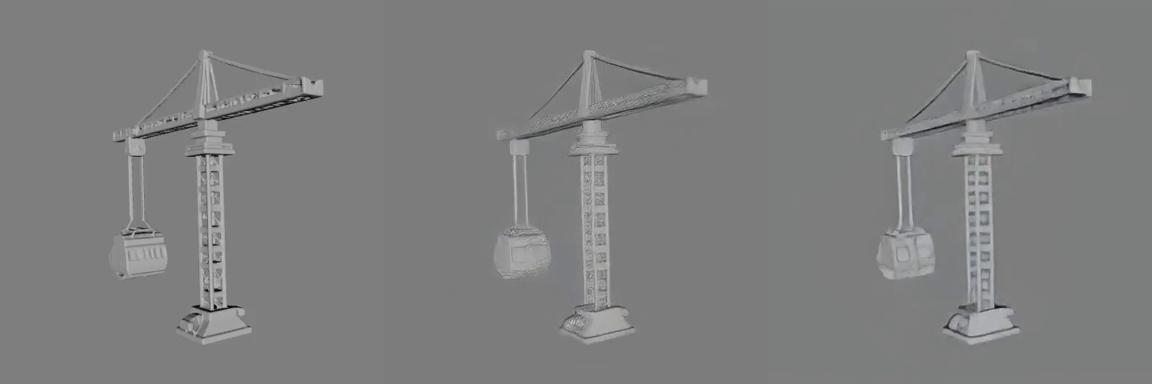}
    \vspace{-1em}
    \caption{Qualitative comparison of the primitive Texture Remover and its distilled variant at 3 denoising steps.}
    \label{fig:distill_comparision}
\end{figure}

\noindent \textbf{The performance of distilled Texture Remover}. We find that CFG is unnecessary for the Texture Remover (as shown in Figure~\ref{fig:tr_cfg} and Table~\ref{tab:original_tr}). Moreover, to further accelerate inference, we reduce the original 50 denoising steps to 3 via distillation, making the Texture Remover fast enough to be integrated into training. Table~\ref{tab:distilled_tr} reports the results of applying DMD2 distillation with different training steps and evaluating the distilled 3-step models. Compared with the original (undistilled) 3-step Texture Remover, the distilled variants consistently achieve better performance. We ultimately select the checkpoint distilled for 300 steps, as it attains the lowest MSE and highest PSNR. Notably, when distillation continues beyond 600 steps, all metrics deteriorate rapidly, indicating overfitting. Figure~\ref{fig:distill_comparision} compares the original Texture Remover and the distilled variant under the same 3-step denoising setting. The original Texture Remover produces noticeably blurred regions, which may interfere with subsequent Refaçade training, whereas the distilled Texture Remover yields cleaner and more reliable results.

\begin{table}[t]
\caption{Ablation study for inference steps and CFG scales of primitive texture remover. The value of \textit{Ewarp} falls within the range of \(1 \times 10^{-3}\). The best results are \textbf{boldfaced}, and the second-best results are \underline{underlined}.}
\scriptsize
\label{tab:original_tr}
\centering
\resizebox{0.95\linewidth}{!}{%
\begin{tabular}{c|ccccccc}
\toprule
Infer Steps & Scale & MSE$\downarrow$ & PSNR$\uparrow$ & SSIM$\uparrow$ & LPIPS$\downarrow$  & EWarp$\downarrow$ & Time(s)$\downarrow$\\
\midrule
3   &1.0 & 22.91 & 35.52 & 0.9719 & 0.0279 & 0.4872 & \textbf{5.7787}\\
10  &1.0 & 21.68 & 35.66 & 0.9729 & 0.0263 & 0.4523 & \underline{9.2262} \\
20  &1.0 & \underline{20.18} & \underline{36.01} & \underline{0.9741} & \underline{0.0239} & \textbf{0.4383} & 18.5865\\
50  &1.0 & \textbf{17.33} & \textbf{36.56} & \textbf{0.9767} & \textbf{0.0250} & \underline{0.4407} & 46.7409\\
50  &1.5 & 57.54 & 31.75 & 0.9674 & 0.0427 & 1.6259 & 93.7643\\
50  &2.0 & 191.51 & 26.47 & 0.9510 & 0.0916 & 4.7785 & 93.7643\\
50  &2.5 & 299.97 & 24.38 & 0.9408 & 0.1103 & 6.4976 & 93.7643\\
\bottomrule
\end{tabular}}
\end{table}

\begin{table}[t]
\setlength{\tabcolsep}{3.5pt}
\caption{Ablation study for distillation steps of texture remover, $\text{inference step} = 3$. The value of \textit{Ewarp} falls within the range of \(1 \times 10^{-3}\). The best results are \textbf{boldfaced}, and the second-best results are \underline{underlined}. }
\scriptsize
\label{tab:distilled_tr}
\centering
\scriptsize
\begin{tabular}{c|ccccc}
\toprule
Distill Steps & MSE$\downarrow$ & PSNR$\uparrow$ & SSIM$\downarrow$ & LPIPS$\uparrow$  & EWarp$\downarrow$ \\
\midrule
100  & \underline{19.41} & 36.04 & 0.9720 & \textbf{0.0300} & \textbf{0.4423} \\
200  & 21.37 & 35.72 & 0.9729 & 0.0341 & \underline{0.4533} \\
300  & \textbf{19.16} & \textbf{36.55} & \underline{0.9730} & 0.0371 & 0.4669 \\
400  & 19.35 & 36.47 & \underline{0.9724} & 0.0381 & 0.4739 \\
500  & 19.51 & 36.40 & 0.9726 & 0.0373 & 0.4678 \\
600  & 20.32 & 36.01 & \textbf{0.9739} & 0.0396 & 0.4983 \\
700  & 25.57 & 31.11 & 0.9718 & \underline{0.0331} & 0.6631 \\
800  & 26.76 & 34.94 & 0.9693 & 0.0359 & 0.7460 \\
\bottomrule
\end{tabular}
\end{table}

\begin{table*}[!htbp]
\centering
\scriptsize
\setlength{\tabcolsep}{6.5pt}
\renewcommand{\arraystretch}{1.25}
\caption{Evaluation on ECSSD image dataset. The LPIPS for background evaluates background perseveration, while LPIPS for foreground evaluates the similarity between reference texture and generated content. CLIP, DINO and Dream are the abbreviations of CLIPScore, DINOScore and DreamSim, respectively. The best results are \textbf{boldfaced}, and the second-best results are \underline{underlined}.}
\begin{tabular}{l|c|cccc|cccc|cc}
\toprule
\multirow{2}{*}{Method} & \multirow{2}{*}{Type} & \multicolumn{4}{c|}{Background} & \multicolumn{4}{c|}{Foreground}& \multicolumn{2}{c}{LLM Evaluation} \\
\cline{3-12}
& & MSE$\downarrow$ & PSNR$\uparrow$ & SSIM$\uparrow$ & LPIPS$\downarrow$
  & CLIP$\uparrow$ & DINO$\uparrow$ & LPIPS$\downarrow$ & Dream$\uparrow$ & GPT-5$\uparrow$ & Gemini$\uparrow$ \\
\toprule
BrushNet~\citep{brushnet_ju2024brushnetplugandplayimageinpainting}
& \multirow{4}{*}{Inpainting}
& 361.22 & 24.63 & 0.8471 & 0.0592
& 0.7086 & 0.2069 & 0.7426 & 0.7358 &2.254 & 1.271\\
ControlNet-Inp~\cite{ControlNetInpaint}
& & 113.14 & 29.80 & 0.8487 & 0.2386
& 0.6901 & 0.1808 & 0.7701 & 0.7077 &1.983 & 0.864\\
Flux-Fill~\cite{flux}
& & 1148.02 & 19.89 & 0.5431 & 0.1699
& 0.7200 & 0.2071 & 0.7190 & 0.7599 & 2.034 & 0.983\\
SD3-Inpaint~\cite{sd3_esser2024scaling}
& & 113.14 & 29.80 & 0.8487 & 0.0416
& 0.6901 & 0.1774 & 0.7701 & 0.7077 & 1.797& 0.932\\
\hline
UltraEdit~\citep{ultraedit_zhao2024ultraedit}
& \multirow{8}{*}{General}
& 62.34 & 32.07 & 0.9049 & 0.0255
& 0.6837 & 0.1708 & 0.7679 & 0.7006 & 2.644&1.763\\
Flux-Kont-I~\citep{blackforestlabs2025fluxkontext}
& & 59.08 & 31.88 & 0.9133 & 0.0367
& \underline{0.7918} & \underline{0.4902} & \underline{0.6418} & \underline{0.8253} &2.407&0.847\\
Flux-Kont-T~\citep{blackforestlabs2025fluxkontext}
& & 1651.74 & 20.02 & 0.5558 & 0.1038
& 0.6789 & 0.1719 & 0.7267 & 0.7029 &2.322&2.102\\
HiDream-E1~\citep{hidreami1technicalreport}
& & 2402.24 & 22.39 & 0.7692 & 0.1282
& 0.7008 & 0.2073 & 0.7326 & 0.7223 &2.542&1.746\\
HQ-Edit~\citep{hq_edit_hui2024hq}
& & 7733.84 & 10.39 & 0.2732 & 0.3621
& 0.7017 & 0.2172 & 0.7461 & 0.7223 &1.288&0.983\\
InsP2P~\citep{instructpix2pix_brooks2023instructpix2pix}
& & 2779.51 & 15.93 & 0.4687 & 0.2177
& 0.6933 & 0.1760 & 0.7340 & 0.7155 &1.881&1.661\\
Qwen-I-Edit~\citep{qwenimage_wu2025qwenimagetechnicalreport}
& & 1596.50 & 20.19 & 0.5489 & 0.1369
& 0.6864 & 0.2156 & 0.7228 & 0.7135 &2.797&\textbf{2.764}\\
\hline
\textbf{Ours(stage1)}
& \multirow{2}{*}{Inpainting}
& \textbf{23.45} & \textbf{37.66} & \textbf{0.9653} & \textbf{0.0095}
& 0.7365 & 0.3177 & 0.6726 & 0.7809 &\textbf{2.864}&\underline{2.740}\\
\textbf{Ours(stage2)}
& & \underline{24.73} & \underline{37.99} & \underline{0.9630} & \underline{0.0101}
& \textbf{0.7934} & \textbf{0.5050} & \textbf{0.6395} & \textbf{0.8407} &\underline{2.831} &\textbf{2.764}\\
\hline
\bottomrule
\end{tabular}
\label{tab:ecssd_all}
\end{table*}

\begin{figure*}[t]
    \centering
    \vspace{1em}
    \begin{subfigure}{\textwidth}
        \centering
        {%
          \fontsize{8pt}{9.6pt}\selectfont
          \begin{tabular}{@{}*{7}{>{\centering\arraybackslash}m{0.12\textwidth}}@{}}
            \minibox{Source} & \minibox{Reference} & \minibox{BrushNet}& \minibox{ControlNet} & 
            \minibox{Flux-Fill}& \minibox{Flux-Fill-I} & \minibox{Flux-Fill-T} \\
          \end{tabular}
        }
        \vspace{2pt}
        \includegraphics[width=\linewidth]{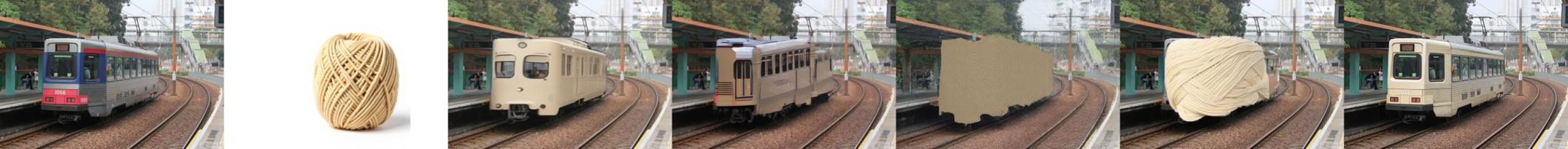}

        \vspace{0pt}

        {%
          \fontsize{8pt}{9.6pt}\selectfont
          \begin{tabular}{@{}*{7}{>{\centering\arraybackslash}m{0.12\textwidth}}@{}}
             \minibox{HiDream-E1}& \minibox{HQ-Edit} & \minibox{InsP2P}& \minibox{Qwen-Image-Edit} & 
             \minibox{SD3-Inpaint}& \minibox{UltraEdit} & \minibox{Ours} \\
          \end{tabular}
        }
        \vspace{2pt}
        \includegraphics[width=\linewidth]{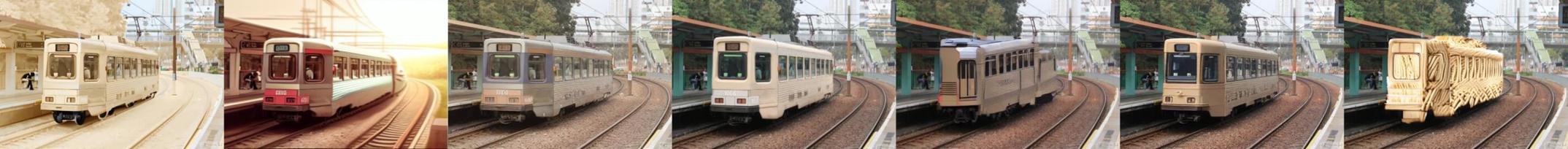}

        \vspace{3pt}
        {\captionsetup{font=footnotesize}
        \fbox{%
          \parbox{\linewidth}{%
            \textbf{Instructive prompt:} \textit{Paint the train in a soft beige color, giving it a smooth, slightly textured fabric-like appearance reminiscent of tightly wound yarn.}\\[2pt]
            \textbf{Descriptive prompt:} \textit{A train in creamy beige, crafted from matte fabric-like material, exudes a cozy aesthetic reminiscent of artisanal craftsmanship.}%
          }%
        }}
    \end{subfigure}

    \vspace{6pt}

    \begin{subfigure}{\textwidth}
        \centering
        {%
          \fontsize{8pt}{9.6pt}\selectfont
          \begin{tabular}{@{}*{7}{>{\centering\arraybackslash}m{0.12\textwidth}}@{}}
            \minibox{Source} & \minibox{Reference} & \minibox{BrushNet}& \minibox{ControlNet} & 
            \minibox{Flux-Fill}& \minibox{Flux-Fill-I} & \minibox{Flux-Fill-T} \\
          \end{tabular}
        }
        \vspace{2pt}
        \includegraphics[width=\linewidth]{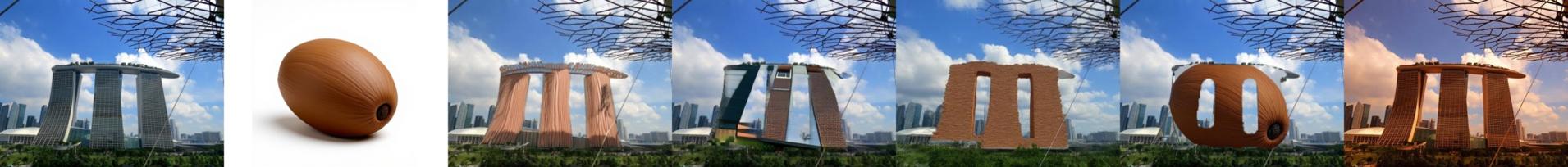}

        \vspace{0pt}

        {%
          \fontsize{8pt}{9.6pt}\selectfont
          \begin{tabular}{@{}*{7}{>{\centering\arraybackslash}m{0.12\textwidth}}@{}}
             \minibox{HiDream-E1}& \minibox{HQ-Edit} & \minibox{InsP2P}& \minibox{Qwen-Image-Edit} & 
             \minibox{SD3-Inpaint}& \minibox{UltraEdit} & \minibox{Ours} \\
          \end{tabular}
        }
        \vspace{2pt}
        \includegraphics[width=\linewidth]{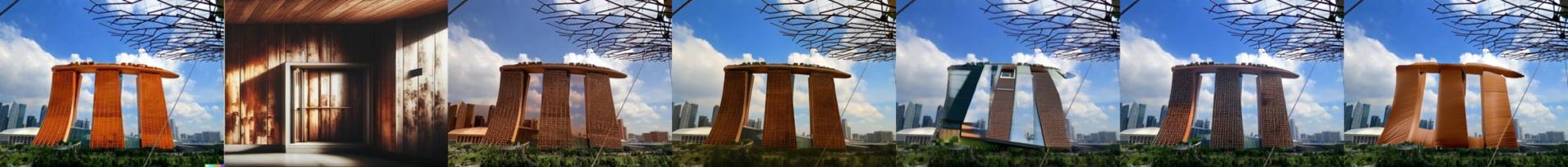}

        \vspace{3pt}
        {\captionsetup{font=footnotesize}
        \fbox{%
          \parbox{\linewidth}{%
            \textbf{Instructive prompt:} \textit{Cover the building in a rich, warm brown tone resembling wood grain texture for a natural and rustic appearance.}\\[2pt]
            \textbf{Descriptive prompt:} \textit{A building in warm terracotta, crafted from smooth, polished clay with subtle wood grain-like textures.}%
          }%
        }}
    \end{subfigure}

    \caption{Qualitative visualization on ECSSD. Each pair of rows uses the instructive and descriptive prompts shown below the images.}
    \label{fig:ecssd_comparision}
\end{figure*}

\begin{table*}[htbp]
\centering
\scriptsize
\setlength{\tabcolsep}{6pt}
\renewcommand{\arraystretch}{1.18}
\caption{Evaluation results on DAVIS dataset. The LPIPS for background evaluates background perseveration, while LPIPS for foreground evaluates the similarity between reference texture and generated content. CLIP, DINO and Dream are the abbreviation of CLIPScore, DINOScore and DreamSim, respectively. Ewarp is at the range of $1\times 10^{-3}$. The best results are \textbf{boldfaced}, the second-best are \underline{underlined}.}
\begin{tabular}{
    l|c|cccc|cccc|c|cc
}
\toprule
\multirow{2}{*}{Method}& \multirow{2}{*}{Type}&\multicolumn{4}{c|}{Background} & \multicolumn{4}{c|}{Foreground} & \multicolumn{1}{c|}{Motion} & \multicolumn{2}{c}{LLM Evaluation} \\ \cline{3-13}
& &{MSE$\downarrow$} & {PSNR$\uparrow$} & {SSIM$\uparrow$} & {LPIPS$\downarrow$}
& {CLIP$\uparrow$} & {DINO$\uparrow$} & {LPIPS$\downarrow$} & {Dream$\uparrow$}
& {EWarp $\downarrow$} & {GPT-5$\uparrow$} & {Gemini$\uparrow$} \\
\midrule
COCOCO ~\citep{cococo_zi2024cococo}
& \multirow{3}{*}{Inpainting}&{3353.94} & {14.10} & {0.5452} & {0.4981}
& {0.6979} & {0.1055} & {0.8076} & {0.6967}
& {11.7707} & 1.644 & 1.644 \\
VACE ~\citep{vace_jiang2025vace}
& &{2175.58} & {15.24} & {0.6199} & {0.3409}
& {0.7182} & {0.1699} & {0.7586} & {0.7141}
& {11.8412} & 2.033 & 2.433 \\
VideoPainter ~\citep{bian2025videopainter}
& &{262.14} & {24.91} & {0.8132} & {0.2425}
& {0.7098} & {0.1573} & {0.7577} & {0.7105}
& {13.6478} & 1.811 & 1.700 \\ \hline
AnyV2V ~\citep{ku2024anyv2v}
& \multirow{9}{*}{General}&1189.29 & 19.00 & 0.6139 & 0.3440
& 0.7182 & 0.1538 & 0.7533 & \underline{0.7317}
& {12.4481} & 2.000 & 2.167 \\
Ditto ~\cite{ditto_bai2025scaling}
& &{1882.43} & {17.51} & {0.6784} & {0.4238}
& {0.6720} & {0.1149} & {0.8118} & {0.6880}
& \underline{9.2721} & 1.300 & 1.333 \\
Flatten~\citep{cong2023flatten}
& &{3662.32} & {13.23} & {0.5597} & {0.5924}
& {0.7231} & {0.1499} & {0.7524} & {0.7380}
& {11.2783} & 1.686 & 1.070 \\
TokenFlow ~\citep{tokenflow_geyer2023tokenflow}
& &{1165.58} & {18.18} & {0.6563} & {0.3972}
& {0.7088} & {0.1311} & {0.7523} & {0.7241}
& {9.2764} & 1.778 & 1.133 \\
ICVE ~\citep{icve_liao2025context}
& &{1513.94} & {18.55} & {0.6501} & {0.4014}
& {0.7018} & {0.1517} & {0.7856} & {0.7158}
& {10.8412} & 1.622 & 1.056 \\
InsV2V ~\cite{insv2v_cheng2024consistent}
& &{3173.99} & {13.77} & {0.5379} & {0.6097}
& {0.6900} & {0.1075} & {0.7602} & {0.7113}
& {12.4250} & 1.822 & 1.422 \\
InsVIE ~\citep{wu2025insvie}
& &{4972.99} & {11.87} & {0.4316} & {0.5970}
& {0.7091} & {0.1447} & {0.8144} & {0.7069}
& {31.3162} & 1.583 & 1.144 \\
Lucy-Edit ~\citep{decartai2025lucyedit}
& &{430.00} & {23.73} & {0.7680} & {0.2708}
& {0.6966} & {0.1563} & {0.7576} & {0.6899}
& {13.2379} & 2.231 & 2.489 \\
Se\~norita~\citep{senorita_zi2025se}
& &{290.22} & {24.56} & {0.7739} & {0.3534}
& {0.6987} & {0.1819} & {0.7456} & {0.6992}
& \textbf{8.6078} & 2.139 & 2.178 \\ \hline
\textbf{Ours(stage1)}
& \multirow{2}{*}{Inpainting} & \underline{51.33} & \underline{32.20} & \underline{0.9160} & \underline{0.0805}
& \underline{0.7183} & \underline{0.2108} & \underline{0.6529} & 0.7269
& 11.1025 & \underline{2.622}& \underline{3.150}\\ 
\textbf{Ours(stage2)} &&\textbf{48.42} & \textbf{32.33} & \textbf{0.9163} & \textbf{0.0795} & \textbf{0.7221} & \textbf{0.2426} & \textbf{0.6373} & \textbf{0.7338} & 10.8550 & \textbf{2.654}& \textbf{3.200} \\ \hline
\bottomrule
\end{tabular}
\vspace{-0.5em}
\label{tab:davis-table}
\end{table*}

\section{Evaluation on Challenging Dataset}

\subsection{Evaluation on Small-resolution Images}

To further investigate the performance of different methods on images, we conduct experiments on the ECSSD~\citep{shi2015hierarchical} dataset. The image resolution in this dataset is relatively low, typically between 200 and 500 pixels on the longer side. We discard samples whose foreground mask area is smaller than 5\% or larger than 90\%.

Table~\ref{tab:ecssd_all} reports the background preservation and foreground texture similarity of all methods. All methods perform inference at the original image resolution. Our approach (Stage~1 and Stage~2) achieves the lowest MSE and LPIPS, together with the highest PSNR and SSIM, indicating the strongest background preservation. Moreover, Stage~2 further improves the retexturing ability over Stage~1, showing higher texture consistency.

In Figure~\ref{fig:ecssd_comparision}, although some methods can roughly turn the train into a beige color, such as Qwen-Image-Edit and Flux-Kontext-Text, they still fail to match the overall color and texture of the reference image. This reflects a fundamental limitation of using text as the sole conditioning signal: even if \textit{beige color} and \textit{fabric-like} are explicitly mentioned, it is difficult to specify the exact RGB values in natural language, and even harder to describe them within a short prompt. Current models also struggle to accurately interpret such RGB-level textual descriptions. In contrast, using a reference image as a conditioning signal offers a clear advantage, as it can provide much richer and more precise information than text alone.

\subsection{Evaluation on Fast-Motion Videos}

Editing fast moving objects in videos is particularly challenging. To evaluate this setting we conduct experiments on the DAVIS~\cite{Perazzi_CVPR_2016} dataset. For a fair comparison we compute all metrics on the first 33 frames of the output videos.

Table~\ref{tab:davis-table} shows that our method achieves state of the art performance on the foreground background and LLM based evaluation metrics, but performs worse in terms of EWarp, indicating slightly reduced temporal consistency. We attribute this limitation to the Texture Remover. Although we synthesize its training set by rendering many pairs of videos with fast moving objects, these data lack nonrigid deformations and only simulate motion through rotations of rigid objects, which introduces a domain gap compared with real world videos. Figure~\ref{fig:davis_compare} provides some visual results.

\section{Limitations and Failure Cases}
\subsection{Limitations}
We believe that the main limitations of our method stem from the Texture Remover. First, the capability of the Texture Remover is entirely inherited from Hunyuan3D~\citep{zhao2025hunyuan3d}. Hunyuan3D is relatively insensitive to textual details, and small characters in particular tend to be treated as texture noise and removed during image-to-mesh reconstruction. This behavior is then learned by the Texture Remover, which causes Refaçade to miss certain fine-grained details. Second, when training the Texture Remover we rely on 3D meshes that are rendered into dynamic videos by translating or rotating the mesh. The reconstructed 3D object is static and cannot deform, which leaves a gap with respect to real-world videos where objects often undergo nonrigid motion. Third, for videos with large motion, some frames may contain motion blur. Such cases are absent from the Texture Remover training data, so the model cannot handle them well, which can lead to structural collapse and chaotic geometry in some Refaçade outputs.

\subsection{Failure Cases}
\begin{figure*}[t]
  \centering

  \begin{subfigure}{0.48\textwidth}
    {\fontsize{8pt}{9.6pt}\selectfont
      \minibox[l]{(a) Object merging introduced by mask dilation.}
    }
    \vspace{2pt}

    \includegraphics[width=\linewidth]{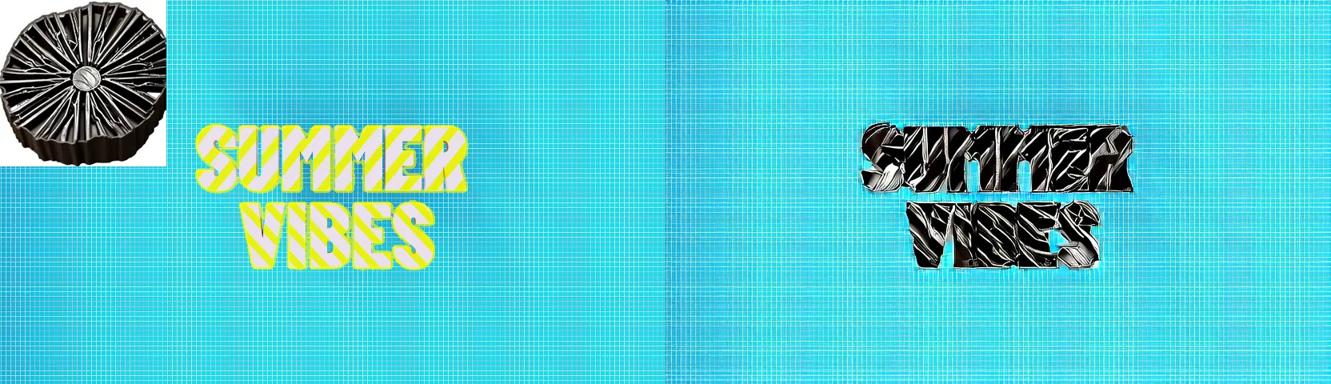}
  \end{subfigure}
  \hfill
  \begin{subfigure}{0.48\textwidth}
    {\fontsize{8pt}{9.6pt}\selectfont
      \minibox[l]{(b) Texture remover struggles with extreme high-frequency details.}
    }
    \vspace{2pt}

    \includegraphics[width=\linewidth]{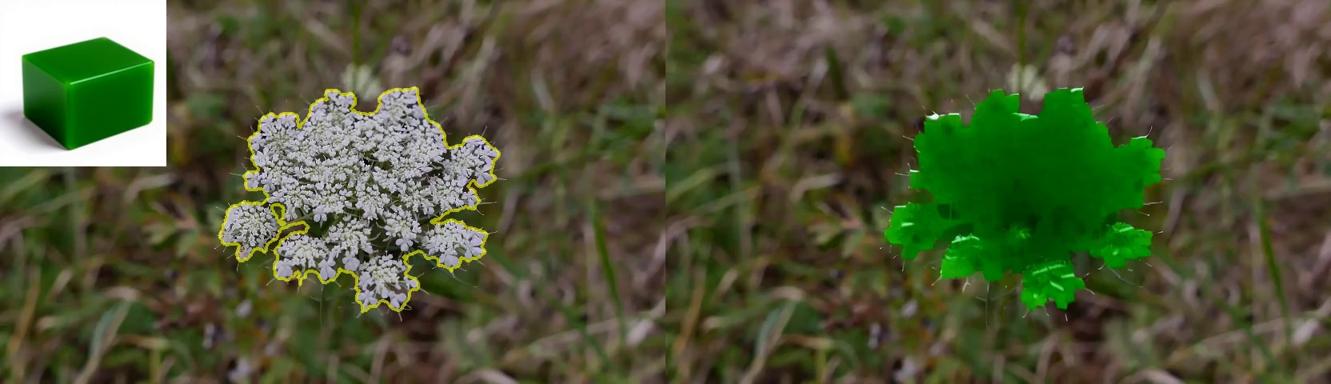}
  \end{subfigure}

  \vspace{6pt}

  \begin{subfigure}{0.48\textwidth}
    {\fontsize{8pt}{9.6pt}\selectfont
      \minibox[l]{(c) Texture remover exhibits low sensitivity to textual information.}
    }
    \vspace{2pt}

    \includegraphics[width=\linewidth]{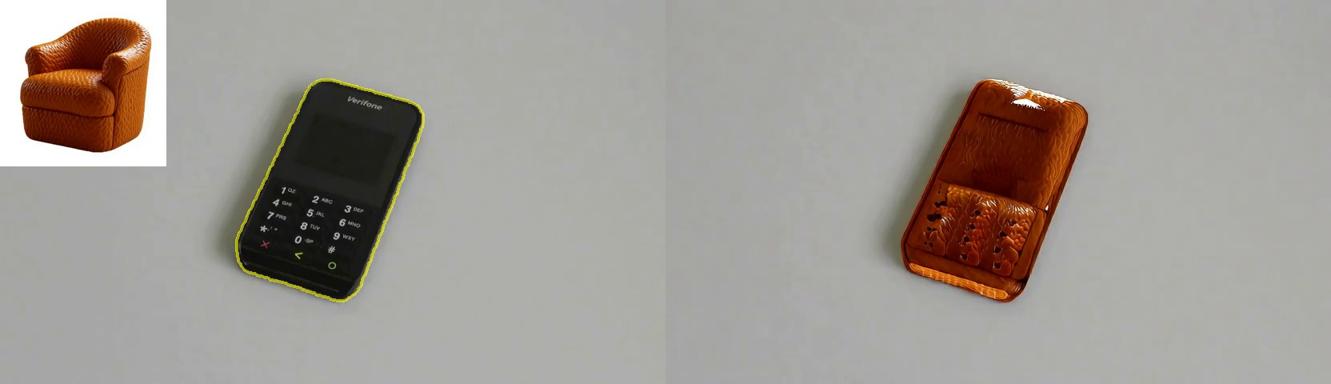}
  \end{subfigure}
  \hfill
  \begin{subfigure}{0.48\textwidth}
    {\fontsize{8pt}{9.6pt}\selectfont
      \minibox[l]{(d) \abbr{} sometimes fails to reshuffle patches properly.}
    }
    \vspace{2pt}

    \includegraphics[width=\linewidth]{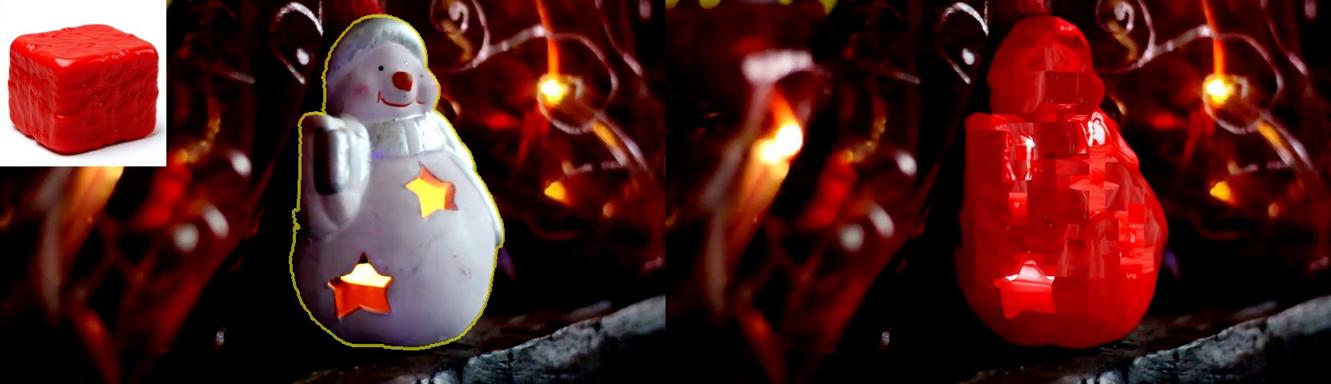}
  \end{subfigure}

  \caption{Failure cases.}
  \label{fig:four_rows_minibox}
\end{figure*}

As shown in Figure \ref{fig:four_rows_minibox}, our model tends to strictly follow the texture of the reference image while overlooking the aesthetic quality of the generated visual content. We attribute this behavior to two factors. First, the classifier free guidance scale used in these examples is suboptimal. Better visual quality can often be obtained by tuning this scale either upward or downward for a given input. Second, scenes that contain untextured objects are more susceptible to reconstruction failures, which in turn lowers the overall success rate of editing.

\section{Future Works}

In future work, we plan to expand the dataset used to train the Texture Remover. Since the current meshes are all rigid and relatively coarse, we aim to incorporate more detailed 4D meshes to enhance the remover’s capability and thereby improve overall robustness. In addition, to further boost the aesthetic quality of \abbr{}, we plan to explore reinforcement learning with reward models.

\begin{figure*}[!htp]
  \centering
  \renewcommand{\arraystretch}{1.0}
  \setlength{\tabcolsep}{0pt}

  \vspace{3pt}

  {%
    \fontsize{8pt}{9.6pt}\selectfont 
    \begin{tabular}{@{}*{8}{>{\centering\arraybackslash}m{0.125\linewidth}}@{}}
      \minibox{Source} & \minibox{Reference} & \minibox{Mask} & \minibox{AnyV2V} &
      \minibox{COCOCO} & \minibox{Ditto} & \minibox{Flatten} & \minibox{ICVE} \\
    \end{tabular}%
  }

  \vspace{-1pt}
  \begin{subfigure}{\linewidth}
    \centering
    \animategraphics[width=\linewidth]{12}{images/davis1/}{1}{24}
  \end{subfigure}

  {%
    \fontsize{8pt}{9.6pt}\selectfont
    \vspace{-4pt}
    \begin{tabular}{@{}*{8}{>{\centering\arraybackslash}m{0.125\linewidth}}@{}}
      \minibox{InsV2V} & \minibox{InsVIE} & \minibox{Lucy-Edit} & \minibox{Señorita} &
      \minibox{Tokenflow} & \minibox{VACE} & \minibox{VideoPainter} & \minibox{Ours} \\
    \end{tabular}%
  }

  \vspace{-1pt}
  \begin{subfigure}{\linewidth}
    \centering
    \animategraphics[width=\linewidth]{12}{images/davis2/}{1}{24}
  \end{subfigure}

  {%
    \fontsize{8pt}{9.6pt}\selectfont 
    \begin{tabular}{@{}*{8}{>{\centering\arraybackslash}m{0.125\linewidth}}@{}}
      \minibox{Source} & \minibox{Reference} & \minibox{Mask} & \minibox{AnyV2V} &
      \minibox{COCOCO} & \minibox{Ditto} & \minibox{Flatten} & \minibox{ICVE} \\
    \end{tabular}%
  }

  \vspace{-1pt}
  \begin{subfigure}{\linewidth}
    \centering
    \animategraphics[width=\linewidth]{12}{images/davis3/}{1}{24}
  \end{subfigure}

  {%
    \fontsize{8pt}{9.6pt}\selectfont
    \vspace{-4pt}
    \begin{tabular}{@{}*{8}{>{\centering\arraybackslash}m{0.125\linewidth}}@{}}
      \minibox{InsV2V} & \minibox{InsVIE} & \minibox{Lucy-Edit} & \minibox{Señorita} &
      \minibox{Tokenflow} & \minibox{VACE} & \minibox{VideoPainter} & \minibox{Ours} \\
    \end{tabular}%
  }

  \vspace{-1pt}
  \begin{subfigure}{\linewidth}
    \centering
    \animategraphics[width=\linewidth]{12}{images/davis4/}{1}{24}
  \end{subfigure}

 {%
    \fontsize{8pt}{9.6pt}\selectfont 
    \begin{tabular}{@{}*{8}{>{\centering\arraybackslash}m{0.125\linewidth}}@{}}
      \minibox{Source} & \minibox{Reference} & \minibox{Mask} & \minibox{AnyV2V} &
      \minibox{COCOCO} & \minibox{Ditto} & \minibox{Flatten} & \minibox{ICVE} \\
    \end{tabular}%
  }

  \vspace{-1pt}
  \begin{subfigure}{\linewidth}
    \centering
    \animategraphics[width=\linewidth]{12}{images/davis5/}{1}{24}
  \end{subfigure}

 {%
    \fontsize{8pt}{9.6pt}\selectfont
    \vspace{-4pt}
    \begin{tabular}{@{}*{8}{>{\centering\arraybackslash}m{0.125\linewidth}}@{}}
      \minibox{InsV2V} & \minibox{InsVIE} & \minibox{Lucy-Edit} & \minibox{Señorita} &
      \minibox{Tokenflow} & \minibox{VACE} & \minibox{VideoPainter} & \minibox{Ours} \\
    \end{tabular}%
  }

  \vspace{-1pt}
  \begin{subfigure}{\linewidth}
    \centering
    \animategraphics[width=\linewidth]{12}{images/davis6/}{1}{24}
  \end{subfigure}

 {%
    \fontsize{8pt}{9.6pt}\selectfont 
    \begin{tabular}{@{}*{8}{>{\centering\arraybackslash}m{0.125\linewidth}}@{}}
      \minibox{Source} & \minibox{Reference} & \minibox{Mask} & \minibox{AnyV2V} &
      \minibox{COCOCO} & \minibox{Ditto} & \minibox{Flatten} & \minibox{ICVE} \\
    \end{tabular}%
  }

  \vspace{-1pt}
  \begin{subfigure}{\linewidth}
    \centering
    \animategraphics[width=\linewidth]{12}{images/davis7/}{1}{24}
  \end{subfigure}

 {%
    \fontsize{8pt}{9.6pt}\selectfont
    \vspace{-4pt}
    \begin{tabular}{@{}*{8}{>{\centering\arraybackslash}m{0.125\linewidth}}@{}}
      \minibox{InsV2V} & \minibox{InsVIE} & \minibox{Lucy-Edit} & \minibox{Señorita} &
      \minibox{Tokenflow} & \minibox{VACE} & \minibox{VideoPainter} & \minibox{Ours} \\
    \end{tabular}%
  }

  \vspace{-1pt}
  \begin{subfigure}{\linewidth}
    \centering
    \animategraphics[width=\linewidth]{12}{images/davis8/}{1}{24}
  \end{subfigure}

  \caption{Comparison results of \abbr{} and baselines on DAVIS. \emph{Best viewed with Adobe Acrobat Reader; click to play.}}
  \label{fig:davis_compare}
\end{figure*}

\clearpage

\begin{figure*}[!htp]
    \centering
    \includegraphics[width=\textwidth]{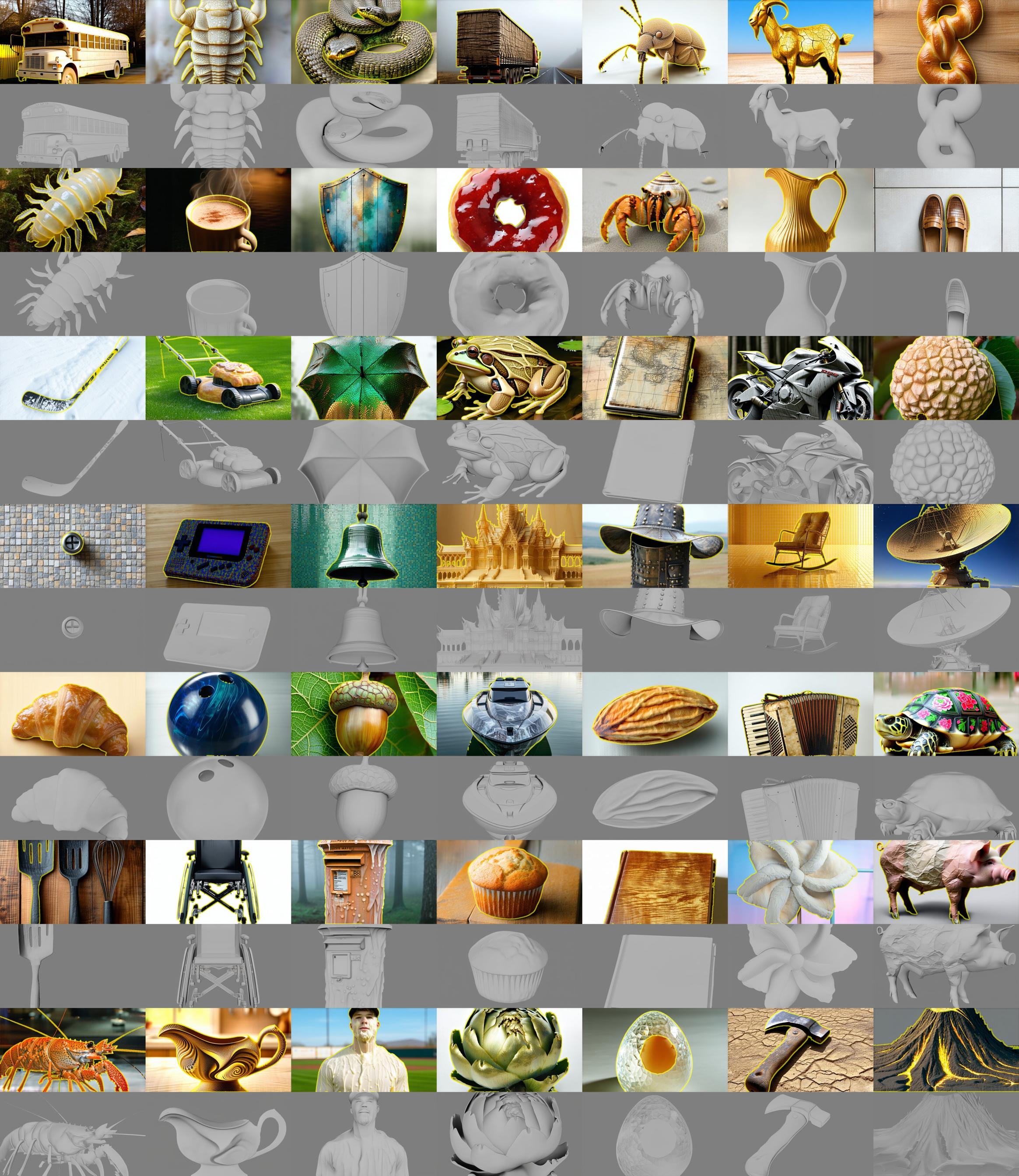}
    \caption{Visual results of Texture Remover.}
    \label{fig:mesh}
\end{figure*}

\begin{figure*}[!htp]
    \centering
    \includegraphics[width=\textwidth]{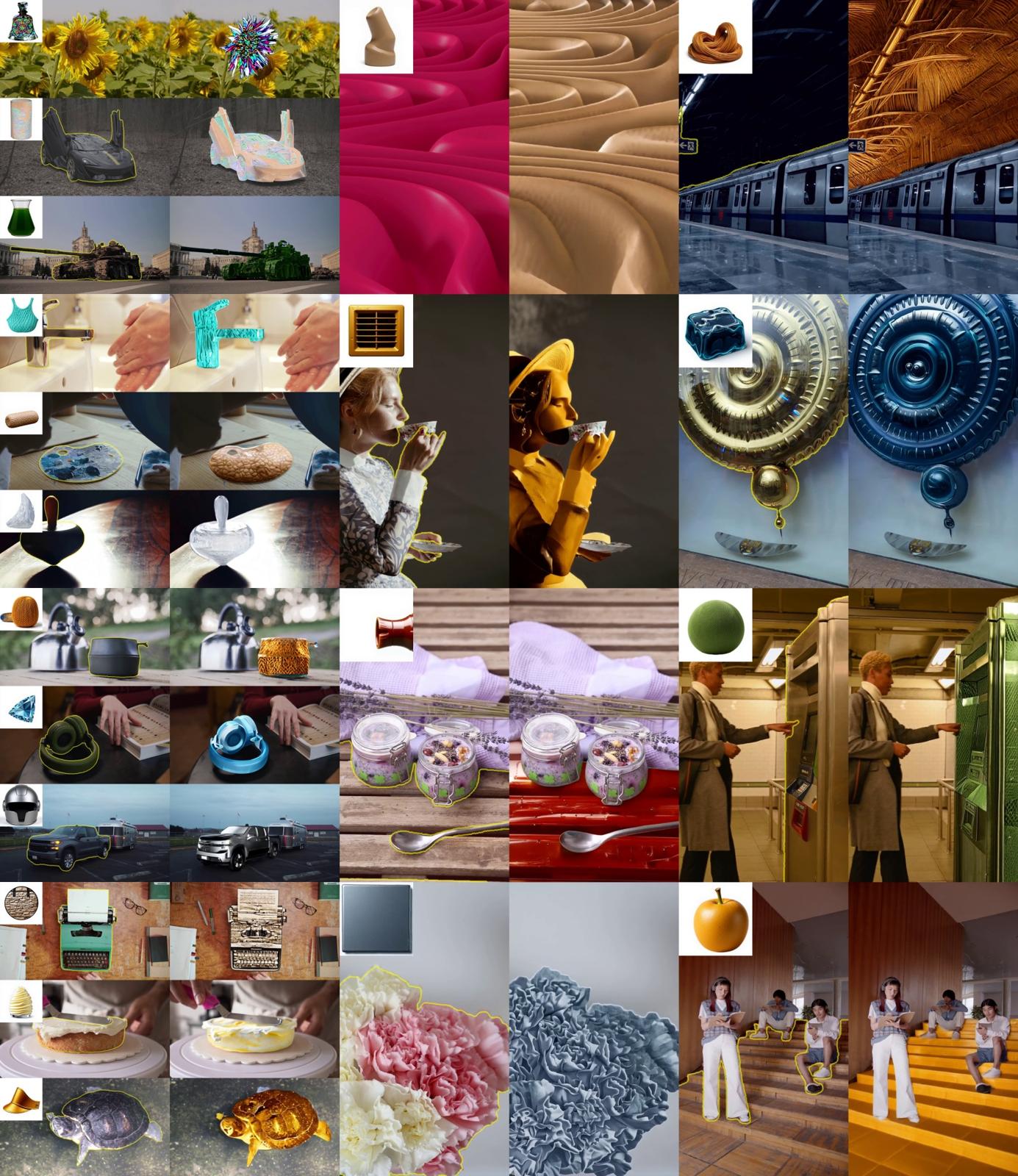}
    \caption{Visual results of \abbr{} on images.}
    \label{fig:appendix_image_results}
\end{figure*}

\begin{figure*}[!htp]
    \centering
    \animategraphics[width=\linewidth]{12}{images/appendix_mp4/}{1}{24}
    \caption{Visual results of \abbr{} on videos. \emph{Best viewed with Adobe Acrobat Reader; click to play.}}
    \label{fig:appendix_video_results}
\end{figure*}

\end{document}